\documentclass[10pt]{scrartcl}

\usepackage{amssymb}
\usepackage{amsmath}
\usepackage{amscd}
\usepackage{amsthm}
\usepackage{graphicx}
\usepackage{caption}
\usepackage{subcaption}
\usepackage{url}

\newtheorem{definition}{Definition}[section]
\newtheorem{theorem}[definition]{Theorem}
\newtheorem{corollary}[definition]{Corollary}
\newtheorem{proposition}[definition]{Proposition}

\renewcommand{\qed}{\hspace*{\fill} $\blacksquare$}

\newcommand{\commentout}[1]{}

\newcommand{\introskip}{\vspace{3ex}}

\newcommand{\R}{\mathbb{R}}                    

\newcommand{\abs}[1]{\mathop{\left\lvert #1 \right\rvert}} 
\newcommand{\args}[1]{\mathop{\left( #1 \right)}} 

\newcommand{\norm}[1]{\mathop{\left\lVert #1 \right\rVert}}
\newcommand{\cbrace}[1]{\mathop{\left\{ #1 \right\}}}
\newcommand{\bracket}[1]{\mathop{\left[ #1 \right]}}

\newcommand{\absS}[2]{\mathop{\left\lvert #1 \right\rvert#2}} 
\newcommand{\argsS}[2]{\mathop{\left( #1 \right)#2}} 

\newcommand{\normS}[2]{\mathop{\left\lVert #1 \right\rVert#2}}

\DeclareMathOperator{\id}{id}                  		

\renewcommand{\S}[1]{{\mathcal{#1}}}           	
\def\vec#1{\mathchoice{\mbox{\boldmath$\displaystyle#1$}}
{\mbox{\boldmath$\textstyle#1$}}
{\mbox{\boldmath$\scriptstyle#1$}}
{\mbox{\boldmath$\scriptscriptstyle#1$}}}

\renewenvironment{cases}{%
\left\{\begin{array}{c@{\quad : \quad}l}}%
{%
\end{array}\right.}

\newcounter{part_counter}
\renewenvironment{part}{
\refstepcounter{part_counter}

\medskip

\noindent
\textbf{\arabic{part_counter}.}%
}%

\begin{document}

\title{Homogeneity of Cluster Ensembles}
\author{Brijnesh J.~Jain \\
 Technische Universit\"at Berlin, Germany\\
 e-mail: brijnesh.jain@gmail.com}
 
\date{}
\maketitle

\begin{abstract} 
The expectation and the mean of partitions generated by a cluster ensemble are not unique in general. This issue poses challenges in statistical inference and cluster stability. In this contribution, we state sufficient conditions for uniqueness of expectation and mean. The proposed conditions show that a unique mean is neither exceptional nor generic. To cope with this issue, we introduce homogeneity as a measure of how likely is a unique mean for a sample of partitions. We show that homogeneity is related to cluster stability. This result points to a possible conflict between cluster stability and diversity in consensus clustering. To assess homogeneity in a practical setting, we propose an efficient way to compute a lower bound of homogeneity. Empirical results using the k-means algorithm suggest that uniqueness of the mean partition is not exceptional for real-world data. Moreover, for samples of high homogeneity, uniqueness can be enforced by increasing the number of data points or by removing outlier partitions. In a broader context, this contribution can be placed as a further step towards a statistical theory of partitions.  
\end{abstract} 

\newpage

\tableofcontents

\newpage

\section{Introduction}

Clustering is a standard technique for exploratory data analysis that finds applications across different disciplines such as computer science, biology, marketing, and social science. The goal of clustering is to group a set of unlabeled data points into several clusters based on some notion of dissimilarity. Inspired by the success of classifier ensembles, consensus clustering has emerged as a research topic \cite{Ghaemi2009,VegaPons2011}. Consensus clustering first generates several partitions of the same dataset. Then it combines the sample partitions to a single consensus partition. The assumption is that a consensus partition better fits to the hidden structure in the data than individual partitions. 

One standard approach of consensus clustering combines the sample partitions to a mean partition \cite{Dimitriadou2002,Domeniconi2009,Filkov2004,Franek2014,Gionis2007,Li2007,Strehl2002,Topchy2005,VegaPons2010}. A mean partition best summarizes the sample partitions with respect to some (dis)similarity function. In general, a mean partition is not unique. Non-uniqueness of a mean partition poses a number of challenges, including (i) comparability, (ii) consistency, (iii) asymptotic behavior, and (iv) clustering stability. 

Uniqueness allows us to directly compare the performance of two clustering ensembles via their respective mean partitions. Without uniqueness, comparing two distributions of partitions based on randomly generated samples can be elusive. Moreover, under reasonable conditions, uniqueness implies strong consistency and gives rise to different versions of the law of large numbers \cite{Jain2015c,Topchy2004}. These findings indicate that without uniqueness, statistical inference based on mean partitions can hardly proceed. With regard to clustering stability, non-uniqueness of the mean partition could potentially entail instability of the clustering.

Despite being a desirable property coming along with several benefits, not much research has been devoted to uniqueness of the mean partition in consensus clustering. In particular, it is unclear under which conditions a sample has a unique mean partition. In addition, it is also unclear whether uniqueness only occurs in trivial and exceptional cases or is a feasible property of practical relevance. To approach these issues, we assume that the set of (hard and soft) partitions is endowed with an intrinsic metric induced by the Euclidean distance.

The contributions of this paper are as follows:
\begin{enumerate}
\item \emph{Conditions of uniqueness}. We establish conditions of uniqueness showing that the mean partition is unique if the cluster ensemble generates sample partitions within a sufficiently small ball. 

\item \emph{Homogeneity}. We propose homogeneity as a measure of how close a sample is to having a unique mean partition. We present a lower-bound of homogeneity that can easily be computed and at the same time identifies outlier-partitions that need to be removed in order to guarantee a subsample with unique mean partition. 

\item \emph{Relationship to cluster stability}. We show that homogeneity of a sample is related to cluster stability. This result points to potentially colliding approaches in clustering: standard clustering advocates stability \cite{AJain2010,Luxburg2010} and consensus clustering advocates diversity \cite{Fern2008,VegaPons2011,Zhou2012}, which will be briefly discussed.

\item \emph{Empirical evidence}.
In experiments we assessed the homogeneity of samples obtained by the k-means algorithm applied to synthetic and real-world data. The results suggest that uniqueness of the mean partition is not exceptional and can be enforced by a larger dataset or by removing outlier-partitions if homogeneity is high. 
\end{enumerate}

Though uniqueness of the mean partition is not a side issue, it is still a strict property not valid for many samples. Homogeneity relaxes this strict property and gives us an alternative way to assess the performance of clusterings and cluster ensembles that goes beyond uniqueness of the mean partition. In a wider context, the results presented in this paper contribute towards a statistical theory of partitions \cite{Jain2015c}.

The rest of this paper is structured as follows: Section \ref{sec:partition-spaces} represents partitions as points of an orbit space. In Section \ref{sec:theory}, we present the theoretical contributions. Section 4 discusses experimental results. Finally, Section \ref{sec:conclusion} concludes with a summary of the main results and with an outlook to further research. Proofs are delegated to the appendix.

\section{Partition Spaces}\label{sec:partition-spaces}

To analyze partitions, we suggest a geometric representation proposed in \cite{Jain2015c}. We first show that a partition can be regarded as a point in some geometric space, called orbit space. Orbit spaces are well explored, possess a rich geometrical structure and have a natural connection to Euclidean spaces \cite{Bredon1972,Jain2015,Ratcliffe2006}. 
Then we endow orbit spaces $\S{P}$ of partitions with a distance function $\delta$ related to the Euclidean metric such that $\args{\S{P}, \delta}$ becomes a geodesic metric space.

\subsection{Partitions}
Let $\S{Z}= \cbrace{z_1, \ldots, z_m}$ be a set of $m$ data points. A partition $X$ of $\S{Z}$ with $\ell$ clusters $\S{C}_1, \ldots, \S{C}_{\ell}$ is specified by a matrix $\vec{X} \in [0,1]^{\ell \times m}$ such that $\vec{X}^T\vec{1}_\ell = \vec{1}_m$, where $\vec{1}_\ell \in \R^\ell$ and $\vec{1}_m \in \R^m$ are vectors of all ones. 

The rows $\vec{x}_{k:}$ of matrix $\vec{X}$ refer to the clusters $\S{C}_k$ of partition $X$. The columns $\vec{x}_{:j}$ of $\vec{X}$ refer to the data points $z_j \in \S{Z}$. The elements $x_{kj}$ of matrix $\vec{X} = (x_{kj})$ represent the degree of membership of data point $z_j$ to cluster $\S{C}_k$. The constraint $\vec{X}^T\vec{1}_\ell = \vec{1}_m$ demands that the membership values $\vec{x}_{:j}$ of data point $z_j$ across all clusters must sum to one.

By $\S{P}_{\ell,m}$ we denote the set of all partitions with $\ell$ clusters over $m$ data points. Since some clusters may be empty, the set $\S{P}_{\ell,m}$ also contains partitions with less than $\ell$ clusters. Thus, we consider $\ell \leq m$ as the maximum number of clusters we encounter. If the exact numbers $\ell$ and $m$ do not matter or are clear from the context, we also write $\S{P}$ for $\S{P}_{\ell, m}$. A hard partition $X$ is a partition with matrix representation $\vec{X} \in \cbrace{0,1}^{\ell \times m}$. The set $\S{P}^+ \subset \S{P}$ denotes the subset of all hard partitions. 

\subsection{Orbit Spaces}

We define the \emph{representation space} $\S{X}$ of the set $\S{P} = \S{P}_{\ell, m}$ of partitions by
\[
\S{X} = \cbrace{\vec{X} \in [0,1]^{\ell \times m} \,:\, \vec{X}^T\vec{1}_\ell = \vec{1}_\ell}.
\]
Then we have a natural projection
\[
\pi: \S{X} \rightarrow \S{P}, \quad \vec{X} \mapsto X = \pi(\vec{X})
\]
that sends matrices $\vec{X}$ to partitions $X$ they represent. The map $\pi$ conveys two properties: (1) each partition can be represented by at least one matrix, and (2) a partition may have several matrix representations. 
 
Suppose that matrix $\vec{X} \in \S{X}$ represents a partition $X \in \S{P}$. The subset of all matrices representing $X$ forms an equivalence class $\bracket{\vec{X}}$ that can be obtained by permuting the rows of matrix $\vec{X}$ in all possible ways. The equivalence class of $\vec{X}$ is of the form
\[
\bracket{\vec{X}} = \cbrace{\vec{PX} \,:\, \vec{P} \in \Pi},
\]
where $\Pi$ is the group of all ($\ell \times \ell$)-permutation matrices. The \emph{orbit space of partitions} is the set 
\[
\S{X}/\Pi = \cbrace{\bracket{\vec{X}} \,:\, \vec{X} \in \S{X}}. 
\]
Informally, the orbit space consists of all equivalence classes $\bracket{\vec{X}}$, we can construct as described above. Mathematically, the orbit space $\S{X}/\Pi$ is the quotient space obtained by the action of the permutation group $\Pi$ on the set $\S{X}$. The equivalence classes $\bracket{\vec{X}}$ are the orbits of $\vec{X}$. The orbits $[\vec{X}]$ are in 1-1-correspondence with the partitions $X = \pi(\vec{X})$. Therefore, we can identify partitions with orbits and $\S{P}$ with $\S{X}/\Pi$. Consequently, we occasionally write $\vec{X} \in X$ if $X = \pi(\vec{X})$. 

\subsection{Intrinsic Metric}

Next, we endow the partition space $\S{P}$ with an intrinsic metric $\delta$ related to the Euclidean distance such that $(\S{P}, \delta)$ becomes a geodesic space. The Euclidean norm for matrices $\vec{X} \in \S{X}$ is defined by
\[
\norm{\vec{X}}= \argsS{\sum_{k = 1}^\ell \sum_{j = 1}^m \absS{x_{kj}}{^2}}{^{1/2}}.
\]
The Euclidean norm induces a distance on $\S{P}$ of the form 
\[
\delta: \S{P} \times \S{P} \rightarrow \R, \quad (X, Y) \mapsto \min \cbrace{\norm{\vec{X} - \vec{Y}}\,:\, \vec{X} \in X, \vec{Y} \in Y}.
\]
Then the pair $\args{\S{P}, \delta}$ is a geodesic metric space \cite{Jain2015c}, Theorem 2.1.

\section{Homogeneity of a Sample}\label{sec:theory}

This section first links consensus clustering to the field of Fr\'echet functions from Mathematical Statistics. Then we present conditions of uniqueness. Based on these conditions, we study how likely is a unique mean partition. For this, we propose homogeneity of a sample as a measure of how close a sample is to having a unique mean. We present an efficient way to compute a lower bound of homogeneity. Finally, this section relates homogeneity to cluster stability and points to potentially conflicting approaches in clustering: stability and diversity in consensus clustering. 

\subsection{Fr\'echet Functions}

In this section, we link the consensus function of the mean partition approach to Fr\'echet functions \cite{Frechet1948}. This link provides access to many results from Statistics in Non-Euclidean spaces \cite{Bhattacharya2012}. 

\introskip

Let $(\S{P}, \delta)$ be a partition space endowed with the metric $\delta$ induced by the Euclidean norm. We assume that $Q$ is a probability measure on $\S{P}$ with support $\S{S}_Q$.\footnote{The support of $Q$ is the smallest closed subset $\S{S}_Q \subseteq \S{P}$ such that $Q(\S{S}_Q) = 1$.} The function 
\[
F_Q: \S{P} \rightarrow \R, \quad Z \mapsto \int_{\S{P}} \delta(X, Z)^2\, dQ(X)
\]
is the expected Fr\'{e}chet function of $Q$. The minimum of $F_Q$ exists but but is not unique, in general \cite{Jain2015c}. Any partition $M\in \S{P}$ that minimizes $F_Q$ is an expected partition. We say $Q$ is homogeneous, if the expected partition of $Q$ is unique. Otherwise, $Q$ is said to be heterogeneous. The minimum $V_Q = F_Q(M)$ is called the variation of $Q$. 

\medskip

Let $\S{S}_Q^n = \S{S}_Q \times \cdots \times \S{S}_Q$ denote the $n$-fold cartesian product of support $\S{S}_Q$. If $\S{S}_n = \args{X_1, X_2, \ldots, X_n} \in \S{S}_Q^n$ is a sample of $n$ partitions, then the (empirical) Fr\'{e}chet function of $\S{S}_n$ is of the form
\begin{align*}
F_n: \S{P} \rightarrow \R, \quad Z \mapsto \frac{1}{n}\sum_{i=1}^n \delta\!\argsS{X_i, Z}{^2}.
\end{align*}
As for expected Fr\'echet functions $F_Q$, the minimum of $F_n$ exists but is not unique, in general \cite{Jain2015c}. Any partition $M\in \S{P}$ that minimizes $F_n$ is a mean partition. The minimum $V_n = F_n(M)$ is the variation of $\S{S}_n$. We say the sample $\S{S}_n$ is homogeneous, if the mean partition of $\S{S}_n$ is unique. Otherwise, $\S{S}_n$ is said to be heterogeneous.

\subsection{Conditions of Uniqueness}

In this section, we show that the expected and mean partition are unique if the partitions to be summarized are contained in a sufficiently small ball. 

\introskip

The ball $\S{B}(Z, r)$ with center $Z \in \S{P}$ and radius $r$ is a set of the form
\[
\S{B}(Z, r) = \cbrace{X \in \S{P} \,:\, \delta(X, Z) \leq r}. 
\]
We call the ball $\S{B}(Z, r)$ homogeneous if there is a bijective isometry 
\[
\psi: \S{B}\!\args{Z, r} \longrightarrow \S{B}\!\args{\vec{Z}, r},
\]
where $\S{B}\!\args{\vec{Z}, r}$ is the ball in the Euclidean space $\S{X}$ centered at representation $\vec{Z} \in Z$. The definition of homogeneous ball is independent of the choice of representation, because two balls in $\S{X}$ at different centers but identical radius $r$ are isometric.

The maximum homogeneity (max-hom) radius $\rho_Z$ at $Z$ is the largest radius for which $\S{B}_Z = \S{B}(Z, \rho_Z)$ is a homogeneous ball. We call $\S{B}_Z$ the max-hom ball centered at $Z$. The next result guarantees uniqueness of the expectation and mean if the partitions to be summarized are contained in an open subset of some max-hom ball.
\begin{theorem}\label{theorem:uniqueness-of-mean}
Let $Q$ be a probability measure on $\S{P}$ with support $\S{S}_Q$. Suppose that there is a partition $Z \in \S{P}$ and an open subset $\S{U} \subset \S{B}_Z$ such that $\S{S}_Q \subseteq \S{U}$. Then $Q$ and any sample $\S{S}_n \in \S{S}_Q^n$ are homogeneous. 
\end{theorem}

Note that Theorem \ref{theorem:uniqueness-of-mean} makes no statement about the existence and size of max-hom balls. Therefore, it is unclear whether the uniqueness conditions are satisfied only in exceptional cases or are of practical relevance. The following treatment is devoted to this issue. 

\subsection{Asymmetric Partitions}

This section sets the stage for understanding how likely and how feasible are unique expectations and mean partitions. To this end, we introduce the notion of asymmetric partition. Based on the notion of asymmetry, we characterize partitions with positive max-hom radius. 

\introskip

Let $\Pi^* = \Pi \setminus \cbrace{\vec{I}}$ denote the subset of ($\ell \times \ell$)-permutation matrices without identity matrix $\vec{I}$. The degree of asymmetry of a partition $Z \in \S{P}$ is defined by 
\[
\alpha_Z = \min \cbrace{\norm{\vec{Z} - \vec{P}\vec{Z}} \,:\, \vec{Z} \in Z \text{ and } \vec{P} \in \Pi^*}.
\]
A partition $Z$ is asymmetric if $\alpha_Z > 0$. If $\alpha_Z = 0$, the partition $Z$ is called symmetric. The next result establishes a relationship between the degree of asymmetry and the max-hom radius. 

\begin{proposition}\label{prop:asymmetry-max-hom} 
Let $Z \in \S{P}$ be a partition. Then we have
\begin{enumerate}
\item $\alpha_Z / 4 \leq \rho_Z$
\item $\alpha_Z > 0 \; \Leftrightarrow \; \rho_Z > 0$.
\end{enumerate}
\end{proposition}

\medskip 

\noindent
Proposition \ref{prop:asymmetry-max-hom}(1) says that $\S{A}_Z = \S{B}(Z, \alpha/4)$ is a homogeneous ball. We call $\S{A}_Z$ the asymmetry ball of $Z$. From Theorem \ref{theorem:uniqueness-of-mean} and $\S{A}_Z \subseteq \S{B}_Z$ follows that expectation and mean are unique if the support $\S{S}_Q$ is contained in an open subset of $\S{A}_Z$. 

Proposition \ref{prop:asymmetry-max-hom}(2) states that being an asymmetric partition and having a positive max-hom radius are equivalent properties. Thus, we can characterize partitions with positive max-hom radius by asymmetric partitions: 
\begin{proposition}\label{prop:properties:asymmetry} \ 
\begin{enumerate}
\item Almost all partitions are asymmetric. 
\item A partition is asymmetric if and only if its clusters are mutually distinct.
\end{enumerate}
\end{proposition}

\medskip

\noindent 
The first assertion of Prop.~\ref{prop:properties:asymmetry} states that partitions with degenerated max-hom ball that collapse to a single point are the pathological cases in the sense that they are contained in some subset of measure zero.

The second assertion of Prop.~\ref{prop:properties:asymmetry} provides us a way of how to compute the degree of asymmetry as we will see shortly. Recall that a cluster of a partition $Z$ is represented by a row $\vec{z}_k$ of a representation $\vec{Z} \in Z$. Empty clusters are represented by zero rows. A pair of clusters of $Z$ is distinct if the corresponding rows of $\vec{Z}$ are distinct. The next results are an immediate consequence of Prop.~\ref{prop:properties:asymmetry}(2). 

\begin{corollary} \ 
\begin{enumerate}
\item Every hard partition with at most one empty cluster is asymmetric.
\item A symmetric partition has at least one pair of identical clusters. 
\item A partition with more than one empty cluster is symmetric.
\end{enumerate}
\end{corollary}

\medskip

\noindent 
Next, we show how the degree of asymmetry of a partition can be determined.
\begin{proposition}\label{prop:properties:alpha} Let $Z \in \S{P}_{m,\ell}$ be a partition. 
\begin{enumerate}
\item Let $\vec{Z} \in Z$ be a representation with rows $\vec{z}_1, \ldots, \vec{z}_\ell$. Then
\begin{align*}
\alpha_Z = \min \cbrace{\sqrt{2}\cdot\norm{\vec{z}_p - \vec{z}_q} \,:\, 1 \leq p < q \leq \ell}
\end{align*}
\item Suppose that $Z$ is a hard partition. Then 
\begin{align*}
\alpha_Z = \sqrt{2\args{m_1 + m_2}},
\end{align*}
where $m_1 \leq m_2$ are the sizes of the two smallest clusters of $Z$.
\item Suppose that $Z$ is an asymmetric hard partition. Then 
\[
\sqrt{2} \leq \alpha_Z \leq 2 \cdot \sqrt{\left\lceil\frac{m}{\ell}\right\rceil}.
\]
where $\lceil x \rceil$ denotes the smallest integer larger than or equal to $x$.
\end{enumerate}
\end{proposition}

\medskip

The first statement of Prop.~\ref{prop:properties:asymmetry} tells us how to compute the degree of asymmetry of an arbitrary partition. The second statement of Prop.~\ref{prop:properties:asymmetry} gives us a simpler formula for computing the degree of asymmetry for the subset of hard partitions. Finally, the last statement of Prop.~\ref{prop:properties:asymmetry} tells us the range of values the degree of asymmetry can take for the subset of asymmetric hard partitions. Hard partitions have largest degree of asymmetry if the data points are evenly distributed across all clusters. Conversely, the degree of asymmetry is small if there are clusters with few data points.

\subsection{Homogeneity}

In this section, we introduce homogeneity as a measure of how close a sample is to having a unique mean and provide a lower bound that can be easily determined.

\introskip

Suppose that $\S{S}_n = \args{X_1, \ldots, X_n} \in \S{S}_Q^n$ is a sample of $n$ partitions. By $\S{H}\args{\S{S}_n}$ we denote the set of all homogeneous sub-samples of $\S{S}_n$, that is the set of all sub-samples of $\S{S}_n$ with unique mean partition. Obviously, $\S{H}\args{\S{S}_n}$ is non-empty, because sub-samples consisting of a singleton are homogeneous. If $\S{S}_n$ is homogeneous, then $\S{H}\args{\S{S}_n}$ coincides with the power set of $\S{S}_n$. 

The homogeneity of a sample $\S{S}_n$ is defined by
\[
H\!\args{\S{S}_n} = \max \cbrace{\frac{\abs{S}}{n} \,:\, \S{S} \in \S{H}\args{\S{S}_n}}.
\]
Homogeneity measures how close a sample is to being homogeneous. Homogeneity quantifies the largest fraction of partitions that have a unique mean partition. Conversely, the value $1-H$ tells us how many partitions we need to remove from $\S{S}_n$ to obtain a sub-sample with unique mean partition. Homogeneous samples have homogeneity one and heterogeneous samples have homogeneity less than one. In the worst case, the homogeneity of a sample $\S{S}_n$ is $H\!\args{\S{S}_n} = 1/n$. 

\medskip

It is unclear how to compute the homogeneity $H\!\args{\S{S}_n}$ efficiently. We therefore present a procedure to determine a lower bound of $H\!\args{\S{S}_n}$ by using the degree of asymmetry. Let $\S{A}_i$ be the asymmetry ball of the $i$-th sample partition $X_i$. By
\[
\mathbb{I}_{\S{A}_i}\!\args{X_j} = \begin{cases}
1 & X_j \in \S{A}_i\\
0 & \text{otherwise},
\end{cases}
\]
we denote the indicator function of $\S{A}_i$. We can evaluate $\mathbb{I}_{\S{A}_i}\!\args{X_j}$ by first determining the degree of asymmetry $\alpha_i$ of $X_i$ according to Prop.~\ref{prop:properties:alpha}. Then we test membership of $X_j$ in $\S{A}_i$ by evaluating the expression
\begin{align*}
 \delta\!\args{X_i, X_j} &\leq \frac{1}{4}\alpha_i.
\end{align*}
The fraction of sample partitions of $\S{S}_n$ that are contained in $\S{A}_i$ is given by
\[
h_i = \frac{1}{n} \sum_{j=1}^n\mathbb{I}_{\S{A}_i}\!\args{X_j}.
\]
Then the approximated homogeneity ($\alpha$-homogeneity) of sample $\S{S}_n$ is defined as
\[
h^*\!\args{\S{S}_n} = \max_i h_i.
\]
Obviously, the $\alpha$-homogeneity $h^*\!\args{\S{S}_n}$ is a lower bound of $H\!\args{\S{S}_n}$. Note that $h^*\!\args{\S{S}_n} = 1$ implies that the sample $\S{S}_n$ is homogeneous and therefore has a unique mean partition. If $h^*\!\args{\S{S}_n} < 1$ no statement can be made about whether $\S{S}_n$ is homogeneous or heterogeneous. In this case,  $h^*\!\args{\S{S}_n}$ measures how likely a unique mean is. 

\subsection{Clustering Stability}

This section links clustering instability to consensus clustering and sketches how homogeneity is related to clustering stability in a simplified setting. Finally, we briefly point to a potential conflict between cluster stability and diversity in consensus clustering.

\subsubsection{Clustering Instability}

Choosing the number $\ell$ of clusters is a persisting model selection problem in clustering. One way to select $\ell$ is based on the concept of clustering stability. 
The intuitive idea behind clustering stability is that a clustering algorithm should produce similar partitions if repeatedly applied to slightly different datasets from the same underlying distribution. 

Here, we assume that $\S{S}_{n, k} = \args{X_1, \ldots, X_n}$ is a sample of $n$ partitions $X_i \in \S{P}_{k, m}$ of (possibly different) datasets of size $m$ with $k$ clusters. Following \cite{Luxburg2010}, model selection in clustering is posed as the problem of minimizing the function
\[
I_{n, k} = \frac{1}{n^2} \sum_{i=1}^n \sum_{j = 1}^n\Delta_k\!\args{X_i, X_j}
\] 
over all numbers $k$ of clusters such that $1 \leq k_{\min} \leq k \leq k_{\max} \leq m$. Then one option to choose the number $\ell$ of clusters is as follows:
\[
\ell = \arg\min_{k} I_{n, k}.
\]
The function $I_{n,k}$ is called cluster instability and measures the average distance between partitions. Another less common interpretation is that cluster instability measures the average variation $F_{n, k}\!\args{X_i}$ of the sample partitions $X_i$ of $\S{S}_{n,k}$, where
\[
F_{n, k}\!\args{X_i} = \frac{1}{n}\sum_{j = 1}^n\Delta_k\!\args{X_i, X_j}
\]
is the Fr\'echet function of $\S{S}_{n,k}$ with respect to the distance $\Delta_k$. Thus, we can equivalently rewrite cluster instability as 
\[
I_{n,k} = \frac{1}{n} \sum_{i=1}^n F_{n,k}\!\args{X_i}.
\]
The last equation links cluster stability to consensus clustering and to Fr\'echet functions. 

\subsubsection{Homogeneity vs.~Stability}
Intuitively, we expect that the average pairwise distance $I_{k,m}$ between partitions and the average distance $F_{n,k}(M_k)$ to a mean partition are correlated if the underlying distance function $\Delta_k$ is well-behaved. If $\Delta_k$ is a metric, we have (see Section \ref{proof:eq:F<I})
\begin{align}\label{eq:F<I}
F_{n,k}\!\args{M_k} \leq I_{n, k},
\end{align}
where $M_k$ is a mean or medoid partition of sample $\S{S}_{n,k}$.\footnote{A medoid is a sample partition $M_k \in \S{S}_{k,m}$ such that $F_{n,k}\!\args{M_k} \leq F_{n,k}\!\args{X_i}$ for all $1 \leq i \leq n$.} These considerations suggest that the variation $F_{n,k}(M_k)$ can serve as an alternative score function for model selection that is related to cluster instability $I_{n,k}$. We choose the number $\ell$ of clusters according to the rule
\begin{align}\label{eq:rule_l}
\ell = \arg\min_{k} F_{n,k}\!\args{M_k}.
\end{align}
To relate homogeneity to cluster stability consider the (non-symmetric) distance 
\begin{align*}
\Delta_k\!\args{X_i, X_j} = 1 - \mathbb{I}_{\S{A}_i}\!\args{X_j}. 
\end{align*}
The Fr\'echet function of $\S{S}_{n,k}$ takes the form 
\[
F_{n,k}(Z) = \frac{1}{n}\sum_{i = 1}^n 1 - \mathbb{I}_{\S{A}_Z}\!\args{X_i} = 1- \frac{1}{n}\sum_{i = 1}^n \mathbb{I}_{\S{A}_Z}\!\args{X_i} = 1-h_Z,
\] 
where $h_Z$ is the fraction of sample partitions of $\S{S}_{n,k}$ that are contained in the asymmetry ball $\S{A}_Z$ of $Z$. Then we can rewrite the Fr\'echet variation $F_{n,k}\!\args{M_k}$ at a medoid $M_k$ by
\[
F_{n,k}\!\args{M_k} = 1 - h_k^*, 
\]
where $h_k^*$ is the $\alpha$-homogeneity of $\S{S}_{n,k}$. Thus, choosing the number $\ell$ of clusters according to Equation \eqref{eq:rule_l} is equivalent to choosing $\ell$ according to
\[
\ell = \arg\max_k h_k^*.
\]
We choose $\ell$ in such a way that uniqueness of the mean partition is most likely. This shows the relationship between uniqueness of the mean partition and cluster stability.

In contrast to cluster instability, $\alpha$-homogeneity measures stability with respect to the size of smallest clusters. To see this, recall that the degree of asymmetry of a partition $Z$ is 
\[
\alpha_Z = \sqrt{2(m_1+m_2)},
\]
where $m_1$ and $m_2$ are the sizes of the two smallest clusters (see Prop.~\ref{prop:properties:alpha}(2)). Then a hard partition $X$ is in the asymmetry ball $\S{A}_Z$ of hard partition $Z$ if both partitions disagree on at most $(m_1+m_2)/4$ data points. This shows that a clustering is as stable as the smallest clusters in its partition.

\subsubsection{Stability vs.~Diversity}

In consensus clustering it is recommended to use diverse partitions to improve the performance \cite{Fern2008,VegaPons2011,Zhou2012}. Following \cite{Fern2008}, diversity is measured in the same way as cluster instability, namely by the sum of pairwise distances between sample partitions. 

Though diversity corresponds to cluster instability, its application in consensus clustering does not contradict the goal of cluster stability per se. What matters -- from the point of view of cluster stability -- is whether the resulting consensus partitions are stable. Since diversity corresponds to low homogeneity, it is unlikely that a sample of diverse partitions has a unique mean. To comply with cluster stability, the question is under which conditions are two different mean partitions similar? To answer this question, we need the following result proved by \cite{Dimitriadou2002}:

\begin{theorem}\label{theorem:nesuco}
Let $M \in \S{P}$ is a mean partition of the sample $\S{S}_n = \args{X_1, \ldots, X_n} \in \S{P}^n$. Then every representation $\vec{M} \in M$ is of the form 
\begin{align*}
\vec{M} = \frac{1}{n} \sum_{i=1}^n \vec{X}_{\!i},
\end{align*}
where $\vec{X}_{\!i} \in X_i$ are in optimal position with $\vec{M}$, that is $\delta(X_i, M) = \norm{\vec{X}_{\!i}-\vec{M}}$.
\end{theorem}

\medskip

Theorem \ref{theorem:nesuco} describes the form of a mean partition in terms of representations of the sample partition. Since every partition has only finitely many different representations, the set of mean partitions of a given sample is finite and therefore discrete. 

Now suppose that the sample $\S{S}_n$ has two different mean partitions $M$ and $M'$. Let $\vec{M} \in M$ and $\vec{M}' \in M'$ be representations of both mean partitions in optimal position. Applying Theorem \ref{theorem:nesuco} gives
\begin{align*}
\delta(M,M') &= \norm{\vec{M}-\vec{M}'}
= \norm{\frac{1}{n}\sum_{i=1}^n \vec{X}_i - \frac{1}{n}\sum_{i=1}^n \vec{X}_i'}
=\frac{1}{n}\norm{\sum_{i=1}^n \args{\vec{X}_i  -\vec{X}_i'}}
\end{align*}
where  $\vec{X}_{\!i},\vec{X}_{\!i}'  \in X_i$ are representations in optimal position with $\vec{M}$ and $\vec{M}'$, respectively. 
Since both means are different, there is a non-empty subset $\S{J} \subseteq \cbrace{1, \ldots, n}$ of indices such that $\vec{X}_j \neq \vec{X}_j'$ for all $j\in \S{J}$. We obtain
\begin{align*}
\delta(M,M') &\leq  \frac{1}{n} \sum_{i=1}^n \norm{\vec{X}_i  -\vec{X}_i'} =  \frac{1}{n} \sum_{j \in \S{J}} \norm{\vec{X}_j  -\vec{X}_j'}.
\end{align*}
Consider the following conditions:
\begin{enumerate}
\item The index set $\S{J}$ is small, that is $\abs{\S{J}} \ll n$,
\item The degree of asymmetry $\alpha_j$ of partitions $X_j$ is low for all $j \in \S{J}$,
\item The distance $\norm{\vec{X}_j  -\vec{X}_j'}$ is close to $\alpha_j$ for all $j \in \S{J}$.
\end{enumerate}
Then two different mean partitions are similar if condition (1) or if both conditions (2) and (3) hold. Under these conditions, cluster stability and diversity in consensus clustering are not conflicting approaches. 

In general, it is not self-evident that different mean partitions of a sample of diverse partitions are similar. Therefore, we point to the possibility that cluster stability and diversity in consensus clustering can be contradictory approaches in achieving the common goal of improved cluster performance.

\section{Experiments}

The goal of this section is to assess the homogeneity of samples obtained by k-means applied to synthetic and real-world data. 

\subsection{Experiments on Synthetic Data}

\begin{figure}[t]
\centering
\includegraphics[width=0.9\textwidth]{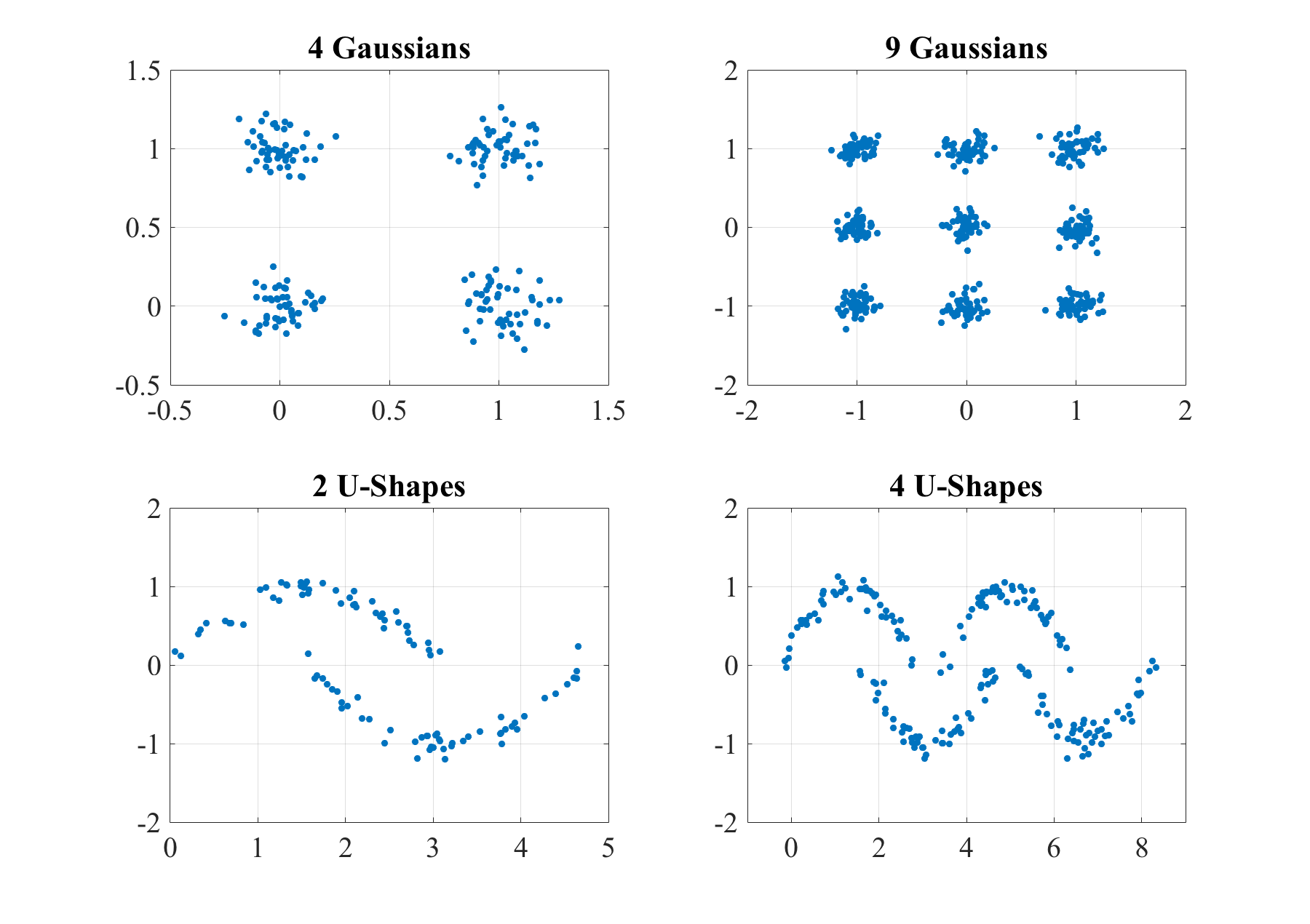}
\caption{Examples of the G4, G9, U2, and U4 dataset.}
\label{datasets.png}
\end{figure}

\paragraph*{Data.} 
We generated the following types of datasets in $\R^2$:

\begin{enumerate}
\itemsep0pt
\item UD: Uniform distribution
\item G4: Four Gaussians 
\item G9: Nine Gaussians
\item U2: Two U-Shapes
\item U4: Four U-Shapes
\end{enumerate}

The UD datasets consists of $m$ data points drawn from the uniform distribution on the unit square.
The G4 and G9 dataset consists of $m$ data points drawn from four and nine Gaussian distributions, resp., with identical covariance matrix $\sigma^2 \vec{I}$ and different mean vectors. The mean vectors of the G2 dataset are the four vertices of the unit square. The nine mean vectors of the G9 dataset are of the form $(x, y)$ with $x, y \in \cbrace{-1, 0, 1}$. 
The U2 and U4 dataset consists of $m$ data points forming concave and convex U-Shapes. The data points were generated by imposing Gaussian noise with mean zero and standard deviation $\sigma$ on the positive and negative component of the sine-function. The U2 dataset has one concave and one convex U-Shape, whereas the U4 dataset has two of both types of U--Shapes. 

Data points were evenly distributed across the different clusters of the G4, G9, U2, and U4 datasets. Figure \ref{datasets.png} shows examples of all datasets with the exception of the UD dataset. The parameters for all datasets were $\sigma = 0.1$ and $m_c = 50$, where $m_c$ is the number of data points of a single cluster (with $m_c = 50$, we have $m_{U\!D} = 50$, $m_{G4} = 200$, $m_{G9} = 450$, $m_{U2} = 100$, and $m_{U4} = 200$).

\paragraph*{Generic Protocol.} A single experiment was conducted according to the following generic scheme:

\medskip

\newcommand{\atab}[1]{\hspace*{#1em}}

\hrule
\vspace{1ex}
\noindent
\emph{Input}:\\
\atab{2} $m$ -- number of data points \\
\atab{2} $\sigma$ \;-- standard deviation \\
\atab{2} $k$ \;-- parameter of k-means

\smallskip

\noindent
\emph{Procedure}:\\
\atab{2} Generate a dataset $\S{Z}$ of size $m$ with standard deviation $\sigma$\\
\atab{2} Repeat $n=100$ times:\\
\atab{3} Apply the k-means algorithm to dataset $\S{Z}$ to obtain sample $\S{S}_{n,k}$ \\
\atab{3} Compute the $\alpha$-homogeneity $h^*\!\args{\S{S}_{n,k}}$

\smallskip

\noindent
\emph{Output}:\\
\atab{2} $\alpha$-homogeneity $h^*\!\args{\S{S}_{n,k}}$
\vspace{1ex}
\hrule

\medskip

The procedure was repeated $100$-times using the same input parameters. Finally, the average  $\alpha$-homogeneity $h^*$ over the $100$ trials was recorded.

\medskip

The UD datasets served as a base-line. For these datasets, $\sigma$ is a factor with which the uniformly generated data points were multiplied.

\subsubsection{Results on G4 Datasets}\label{subsubsec:exp:G4}

The goal of the first series of experiments is to assess homogeneity as a function of the parameters $k$, $\sigma$, and $m$ under the assumption that the cluster structure in the data can be essentially discovered by the k-means algorithm for a suitable value of $k$.

We considered G4 datasets and contrasted the results to those obtained on UD datasets. Unless otherwise stated, the default input parameters were 
\begin{itemize}
\itemsep0pt
\item
$k=4$ for the k-means algorithm,
\item 
$m = 100$ for the size of the datasets,
\item 
$\sigma = 1$ as factor for the UD datasets. 
\end{itemize}

\paragraph*{Homogeneity as a function of $k$.}
We considered three types of datasets: (i) G4 generated with standard deviation $\sigma = 0.05$, (ii) G4 generated with standard deviation $\sigma = 0.7$, and (iii) UD generated with factor $\sigma = 1$. For every $k \in \cbrace{2, \ldots, 10}$ and for all three types of datasets, we conducted experiments according to the above described generic protocol.

\begin{figure}
\centering
\begin{subfigure}{0.495\textwidth}
\centering
\includegraphics[width=\textwidth]{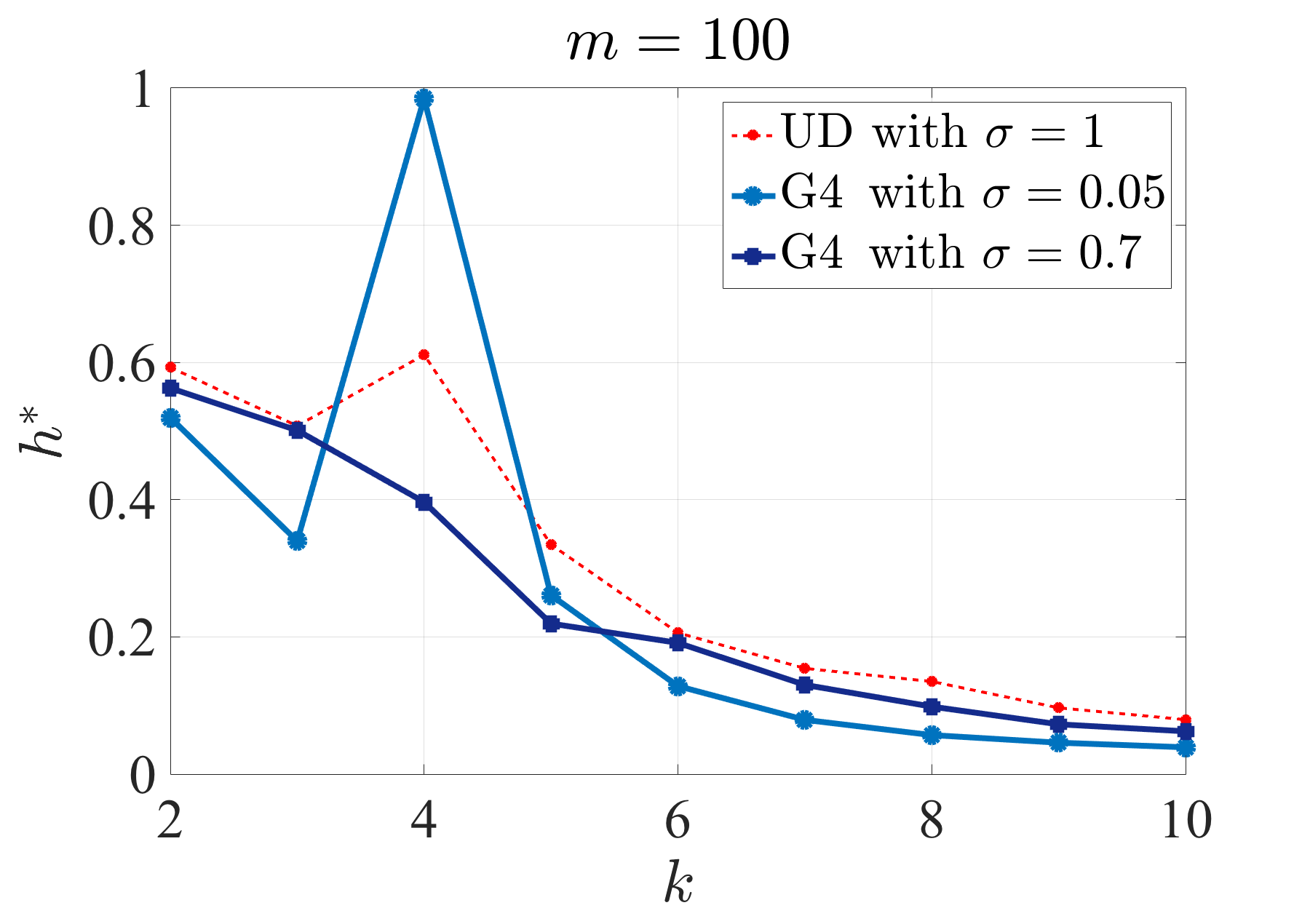}
\caption{Homogeneity as a function of $k$}
\label{fig:g4_k}
\end{subfigure}
\begin{subfigure}{0.495\textwidth}
\centering
\includegraphics[width=\textwidth]{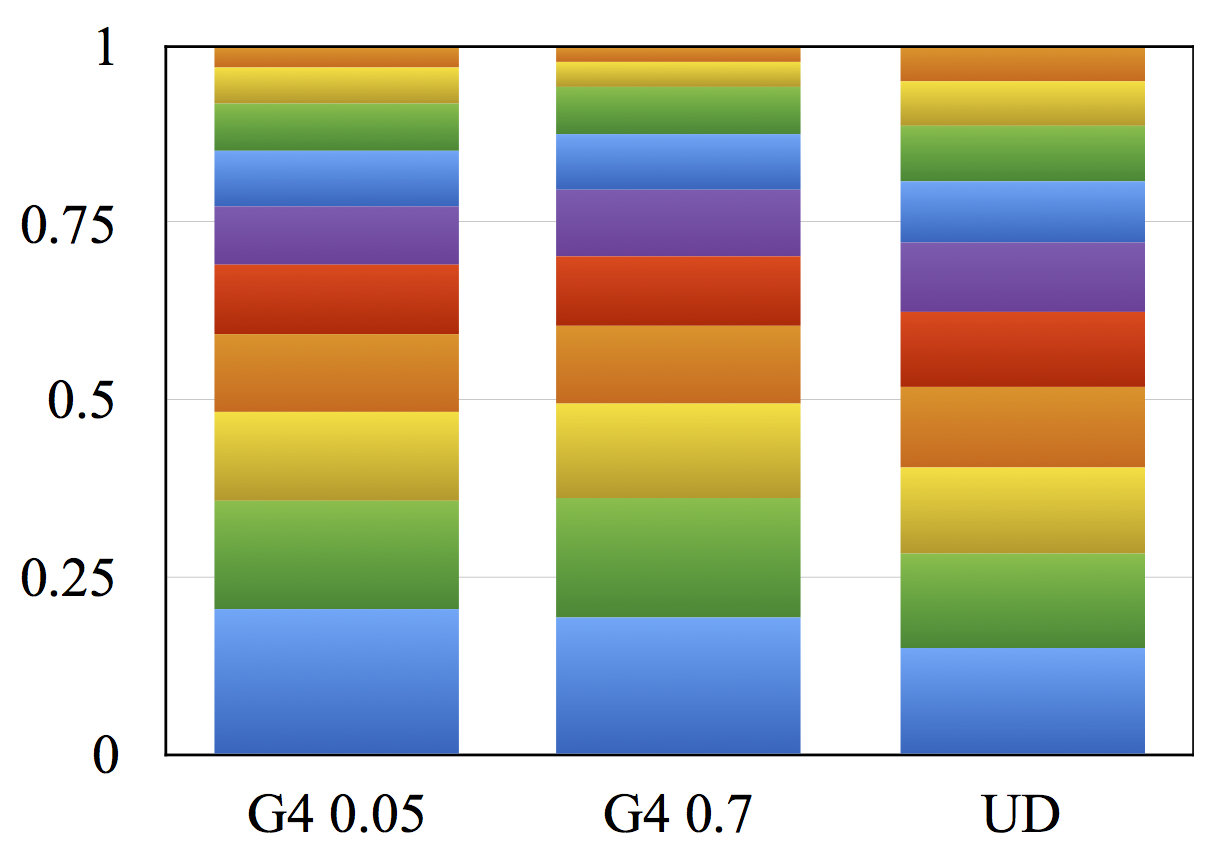}
\caption{Normalized cluster sizes for $k = 10$}
\label{fig:g4_distr}
\end{subfigure}
\caption{Average $\alpha$-homogeneities $h^*$ as a function of the k-means parameter $k$ and normalized cluster sizes for $k=10$ ordered from largest (bottom) to smallest (top), respectively.}
\end{figure}

\medskip

Figure \ref{fig:g4_k} shows the average $\alpha$-homogeneities $h^*$ as a function of the number $k$. We made the following observations:
\setcounter{part_counter}{0}
\begin{part}
The general trend is that homogeneity  decreases with increasing $k$. 
To understand why homogeneity decreases with increasing $k$, recall that the degree of asymmetry of a partition $Z$ is 
\[
\alpha_Z = \sqrt{2(m_1+m_2)},
\]
where $m_1$ and $m_2$ are the sizes of the two smallest clusters (see Prop.~\ref{prop:properties:alpha}(2)). From the strong form of the pigeonhole principle follows 
\[
\frac{m_1+m_2}{2} \leq \frac{m}{k}.
\]
Thus, the sum $m_1+m_2$ decreases with increasing number $k$ of clusters. This means that homogeneity is likely to be lower for large $k$ given a fixed number $m$ of data points. Imbalanced cluster sizes further deteriorate the situation. Figure \ref{fig:g4_distr} shows the average cluster sizes of k-means with $k = 10$ for all three types of datasets. The cluster sizes are normalized by $m$. We see that the cluster sizes of all types of datasets are imbalanced, in particular both types of the G4 datasets.
These findings indicate that increasing the parameter $k$ results in increasingly less stable clusterings and makes a unique mean partition increasingly less likely.
\end{part}

\begin{figure}[t]
\centering
\includegraphics[width=0.97\textwidth]{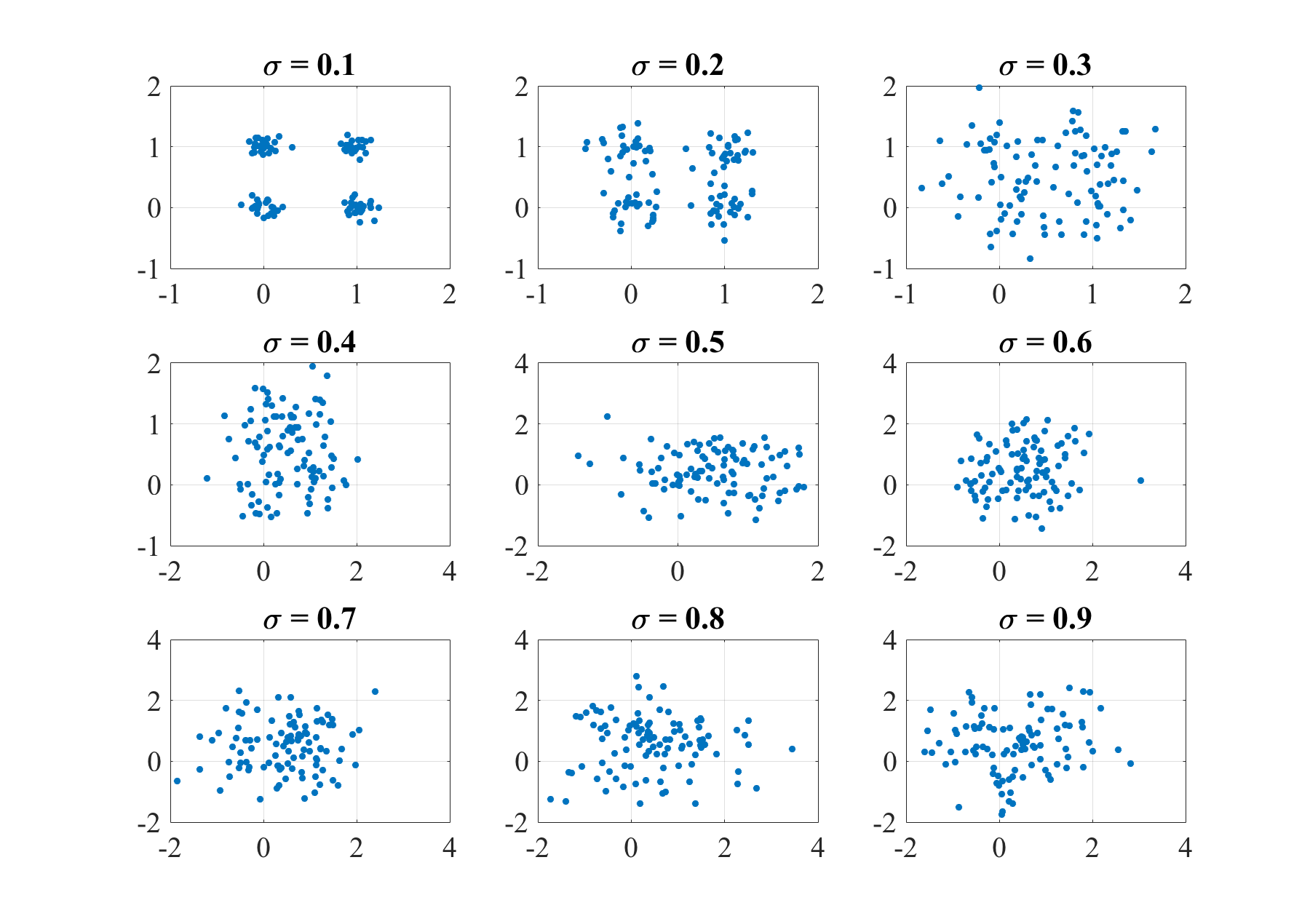}
\caption{Each plot shows $100$ data points randomly generated by four Gaussian distributions with identical mean and increasing standard deviation $\sigma$.}
\label{fig:data4}
\end{figure}

\begin{part} The trend is interrupted at $k=4$ for UD datasets and for G4 datasets with $\sigma = 0.05$.
For $\sigma = 0.05$ the G4 dataset has a clearly visible cluster structure (see Figure \ref{fig:data4}). The k-means algorithm with $k = 4$ recovers this structure resulting in high homogeneity. Only a few partitions need to be removed in order to guarantee uniqueness of the mean.  For $\sigma = 0.7$ no cluster structure is visible as shown in Figure \ref{fig:data4}. Consequently, nothing unexpected happened and homogeneity is low. These findings suggest that uniqueness of the mean partition is more likely when k-means is capable to essentially discover the cluster structure of the dataset.

Surprisingly,  there is a moderate peak at $k=4$ on UD datasets conveying that k-means is most stable when viewing uniformly distributed data as a $2$, and $4$ cluster problem (homogeneity of samples of partitions with one cluster is always one). This moderate peak is even more notable when compared to the results on G4 datasets with $\sigma = 0.7$. This result indicates that peaks in homogeneity do not necessarily allow us to draw conclusions about the cluster structure in a dataset.

By combining both findings, we hypothesize that an evident cluster structure discovered by the underlying algorithm implies peaks in homogeneity but the converse claim does not necessarily hold.
\end{part}

\begin{part} For $k \neq 4$ homogeneity is lower the more structure we assume in the dataset. We assume higher structure in G4 datasets with lower variance and we assume higher structure in any G4 datasets than in UD datasets. The results show that the more evident the assumed cluster structure is the lower is the homogeneity for a wrong choice of $k$. These findings suggest that mismatching an evident cluster structure in the data can introduce additional instability into the clusterings. Further research is necessary to test this hypothesis.
\end{part}

\begin{figure}[t]
\centering
\includegraphics[width=0.5\textwidth]{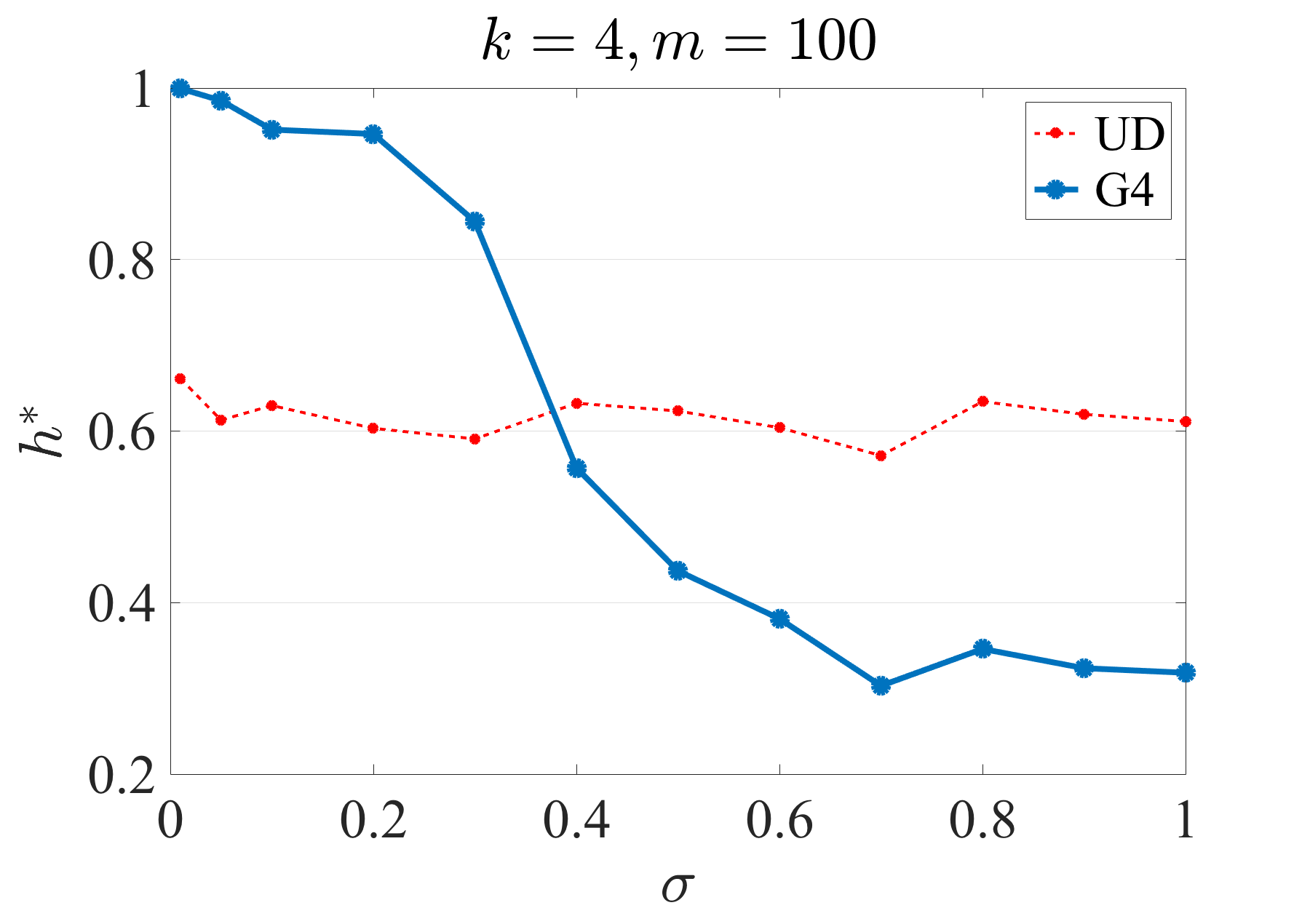}
\caption{Average $\alpha$-homogeneity as a function of $\sigma$.}
\label{fig:std4}
\end{figure}

\paragraph*{Homogeneity as a function of $\sigma$.} For every $\sigma \in \cbrace{0.01, 0.05, 0.1, 0.2, \ldots, 1.0}$ and for both datasets of type G4 and UD, we conducted experiments according to the above described  generic protocol. Figure \ref{fig:data4} depicts examples of G4 datasets with varying standard deviation $\sigma$. 

\medskip

Figure \ref{fig:std4} shows the average $\alpha$-homogeneity $h^*$ as a function of the standard deviation $\sigma$. We observed that the average $\alpha$-homogeneity on G4 datasets decreases with increasing standard deviation $\sigma$ until saturation. Homogeneity is at a high level with values above $0.9$, when the four clusters are clearly visible. For larger standard deviations, the clusters become increasingly blurred and the average $\alpha$-homogeneity rapidly drops below $0.4$. The turning point between high and low homogeneity is around $\sigma = 0.3$ and roughly corresponds to the subjective turning point of what we might perceive as a visible cluster structure (cf.~Figure \ref{fig:data4}). Moreover, homogeneity on G4 datasets around the turning point is comparable to homogeneity on UD datasets, which is invariant under scaling of $\sigma$.
Based on these results we raise the hypothesis that an algorithm that essentially discovers a visible cluster structure in the data is stable and guarantees a unique mean partition after removing a small fraction of partitions.

\paragraph*{Homogeneity as a function of $m$.}
We considered three types of datasets: (i) G4 with $\sigma = 0.3$, (ii) G4 with $\sigma = 0.7$, and (iii) $UD$ with factor $\sigma = 1$. Let $m_c$ denote the number of data points of component $c$ of a dataset. Then G4 datasets have size $m = 4m_c$ and UD datasets are of size $m = m_c$. For every 
\[
m_c \in \cbrace{25, 50, 100, 250, 500, 750, 1000, 1500, 2000, 3000, 4000, 5000}
\]
and for all three types of datasets, we conducted experiments according to the above described generic protocol.

\begin{figure}
\centering
\includegraphics[width=0.45\textwidth]{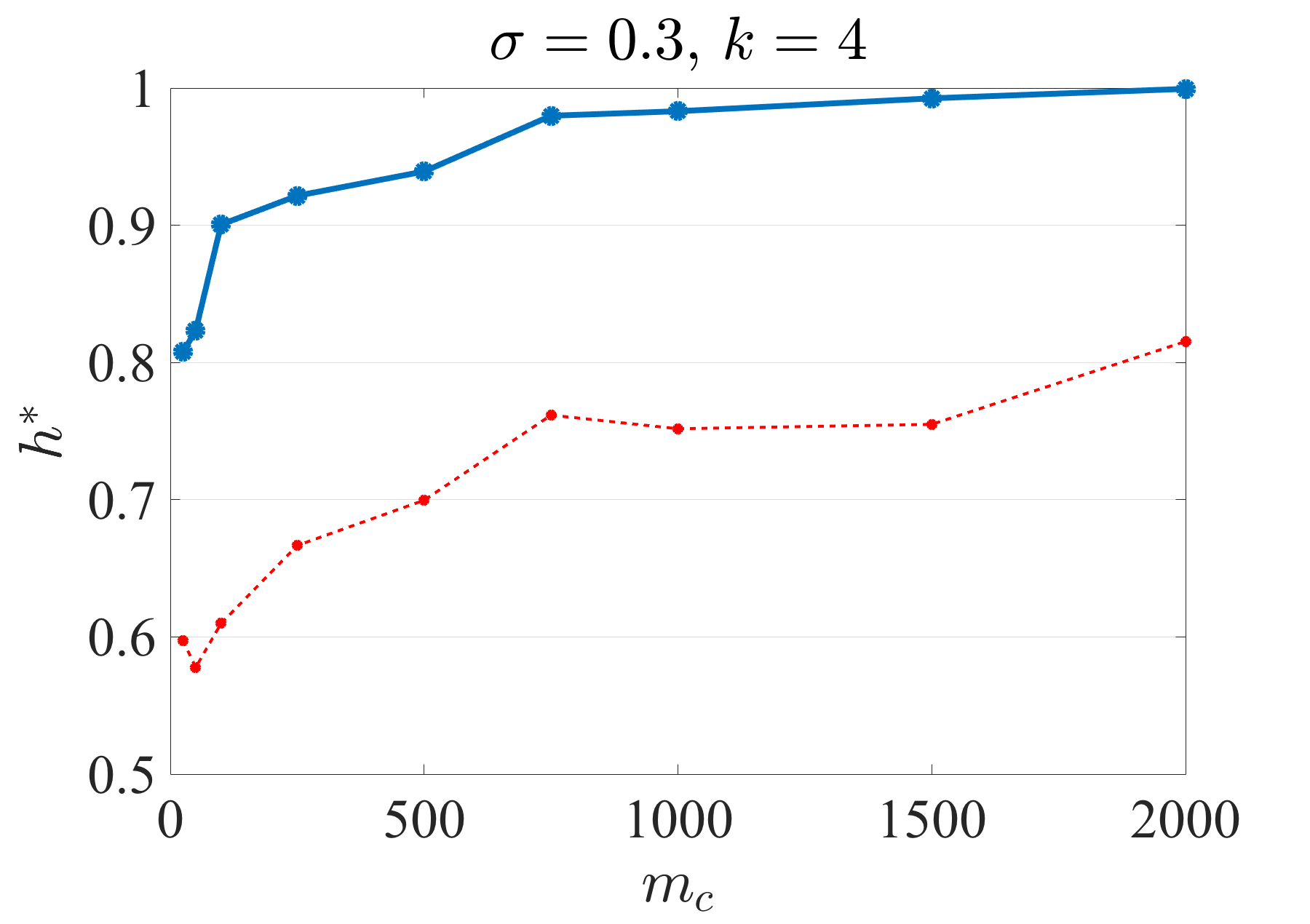}
\includegraphics[width=0.45\textwidth]{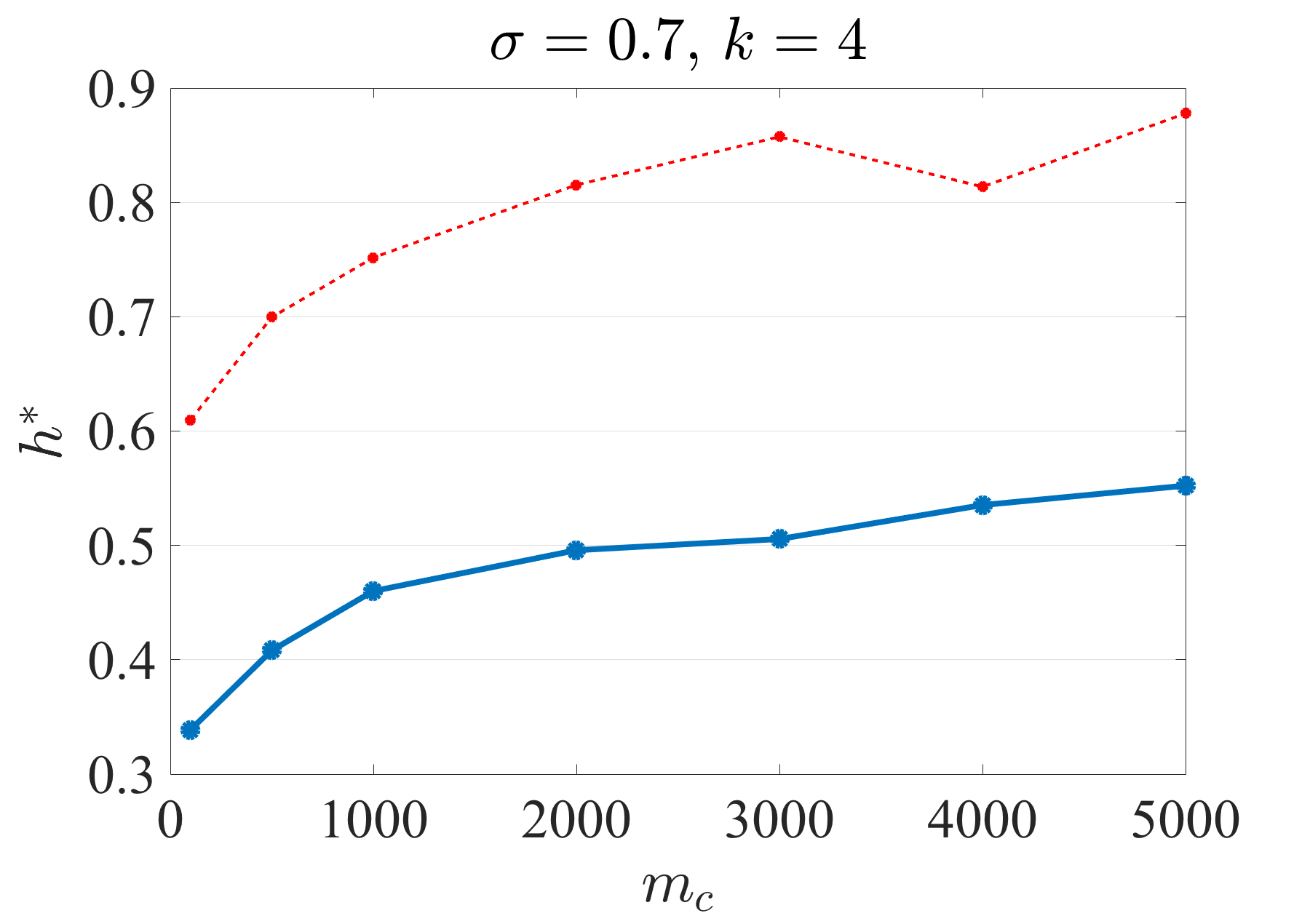}
\caption{Average $\alpha$-homogeneity as a function of the number $m_c$ of data points per cluster. The blue (red) line refers to results on the G4 (UD) datasets.}
\label{m4.png}
\end{figure}

Figure \ref{m4.png} shows an excerpt of the average $\alpha$-homogeneities $h^*$ as a function of the number $m_c$ of data points in each component. We observed that homogeneity increases with increasing dataset size $m$. Moreover, homogeneity increases faster and more substantially for G4 datasets with lower standard deviation. If the four clusters are just visible as for $\sigma = 0.3$, the mean partition is likely to be unique for datasets with at least $m_c = 2000$ data points in each cluster. Surprisingly, homogeneity is high even for UD datasets with more than $5,000$ data points. These results indicate that homogeneity can be improved and uniqueness of the mean partition can be eventually enforced by increasing the size of the dataset.

\subsubsection{Results on U-Shapes and Gaussians}

The goal of the second series of experiments is to assess homogeneity as a function of the parameter $k$ under the assumption that k-means is unable to essentially discover a visible cluster structure in the data. For this we considered U2 and U4 datasets and contrasted the results to those obtained on G4 and G9 datasets. The parameters for all datasets were $\sigma = 0.1$ and $m_c = 50$, where $m_c$ is the number of data points of a single cluster. Figure \ref{datasets.png} shows examples of all four datasets. 

\introskip

\begin{figure}
\centering
\includegraphics[width=0.9\textwidth]{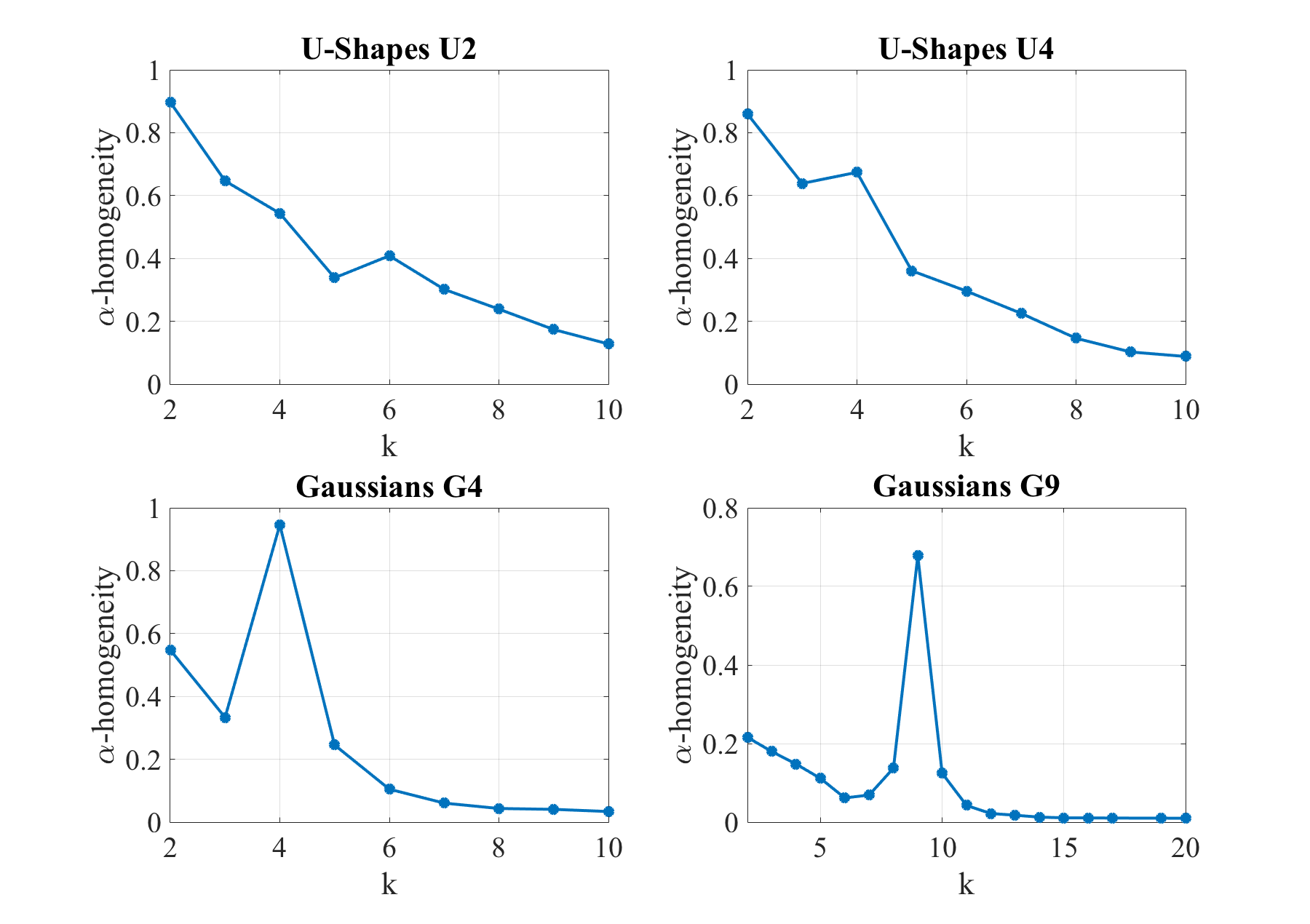}
\caption{Average $\alpha$-homogeneity as a function of the parameter $k$.}
\label{alldata_k.png}
\end{figure}

Figure \ref{alldata_k.png} shows the average $\alpha$-homogeneities for all four datasets. As before, the general trend is that homogeneity decreases with increasing $k$ and is only interrupted when the clearly visible cluster structure in the dataset can be discovered by the k-means algorithm. The peaks at $k=4$ and $k=9$ are evident for the G4 and G9 datasets, respectively. This shows that k-means is most stable and a unique mean partition is most likely when the parameter $k$ coincides with the number of clusters in the datasets. 

\begin{figure}
\centering
\includegraphics[width=0.48\textwidth]{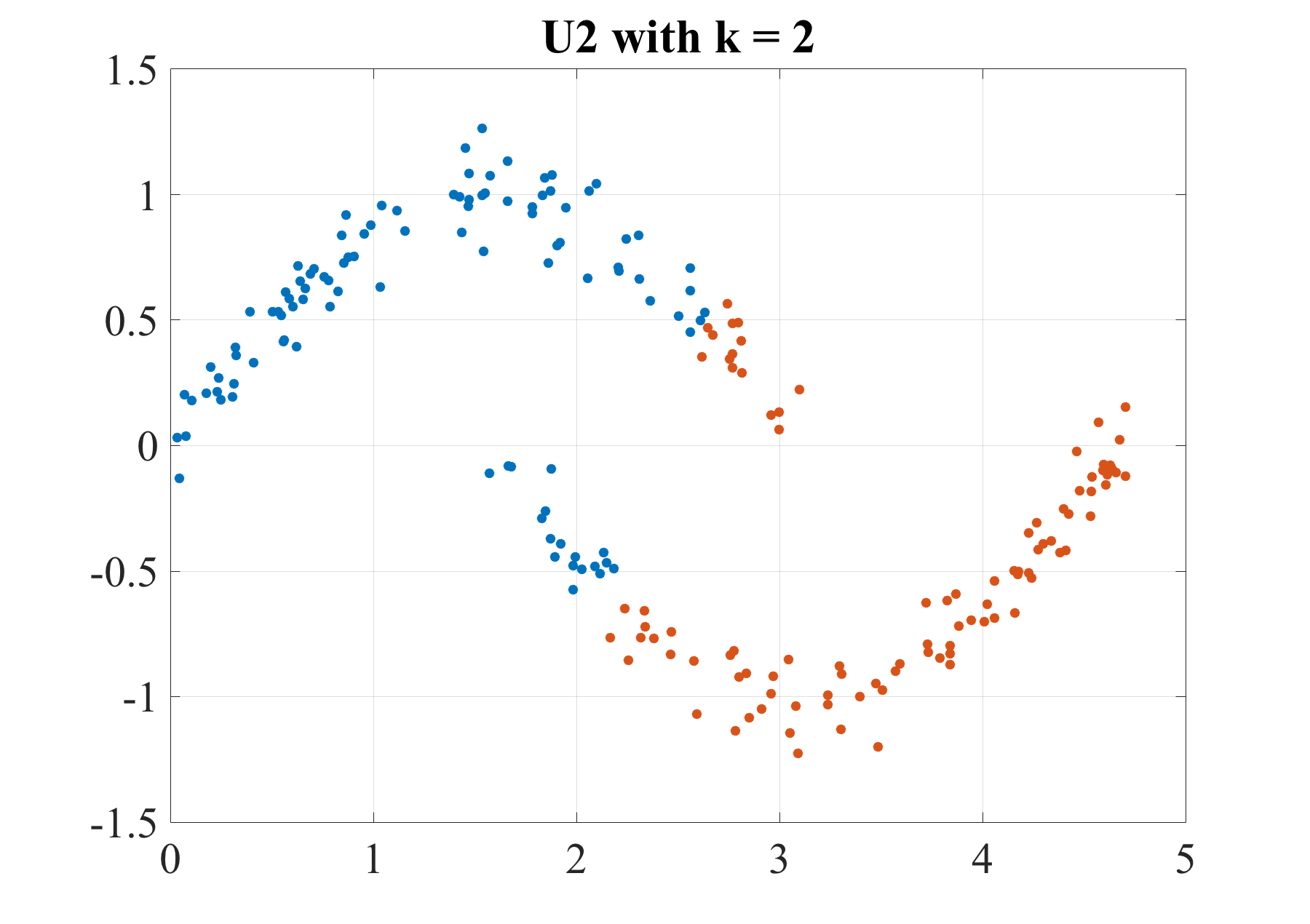}
\includegraphics[width=0.48\textwidth]{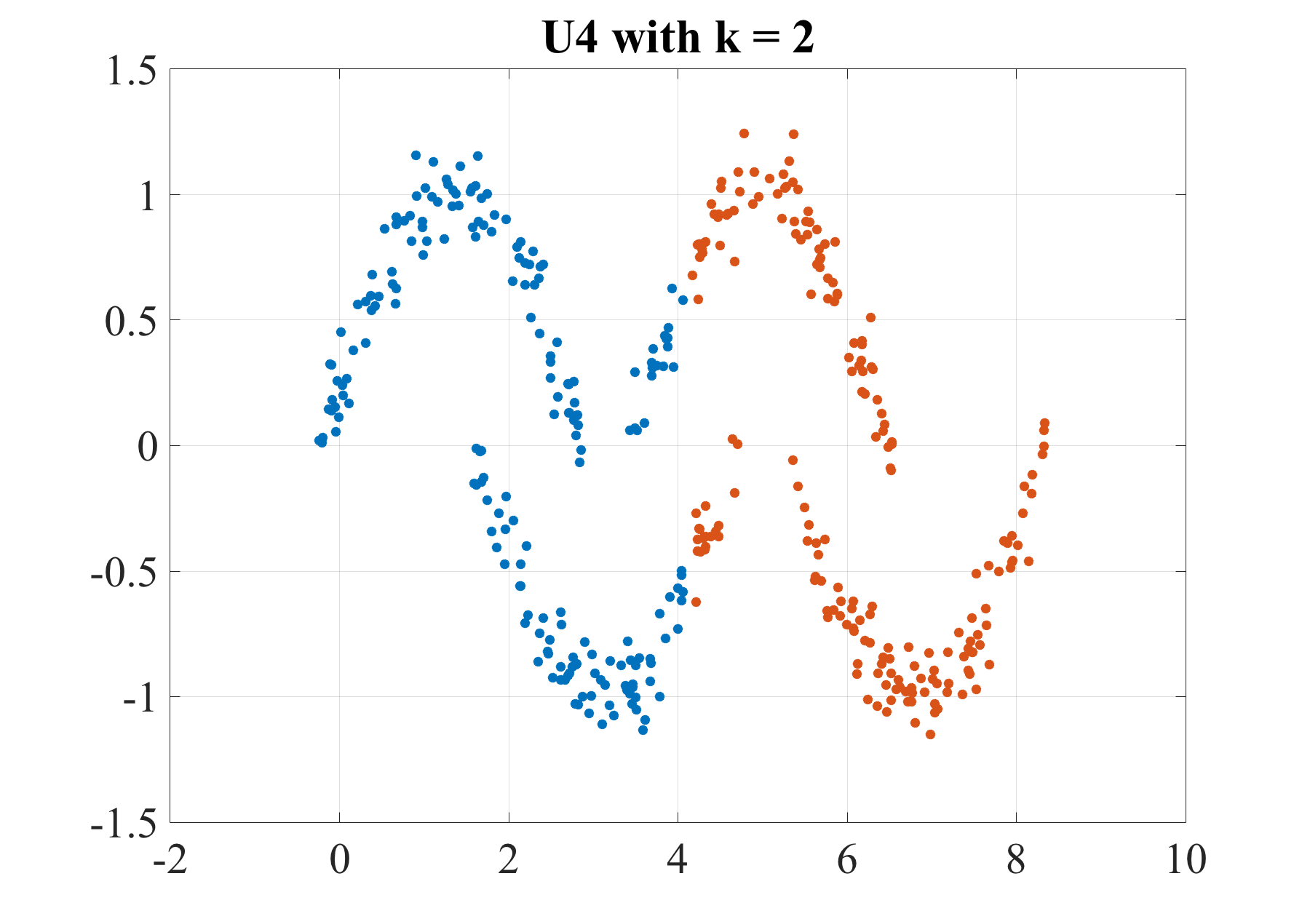}
\includegraphics[width=0.48\textwidth]{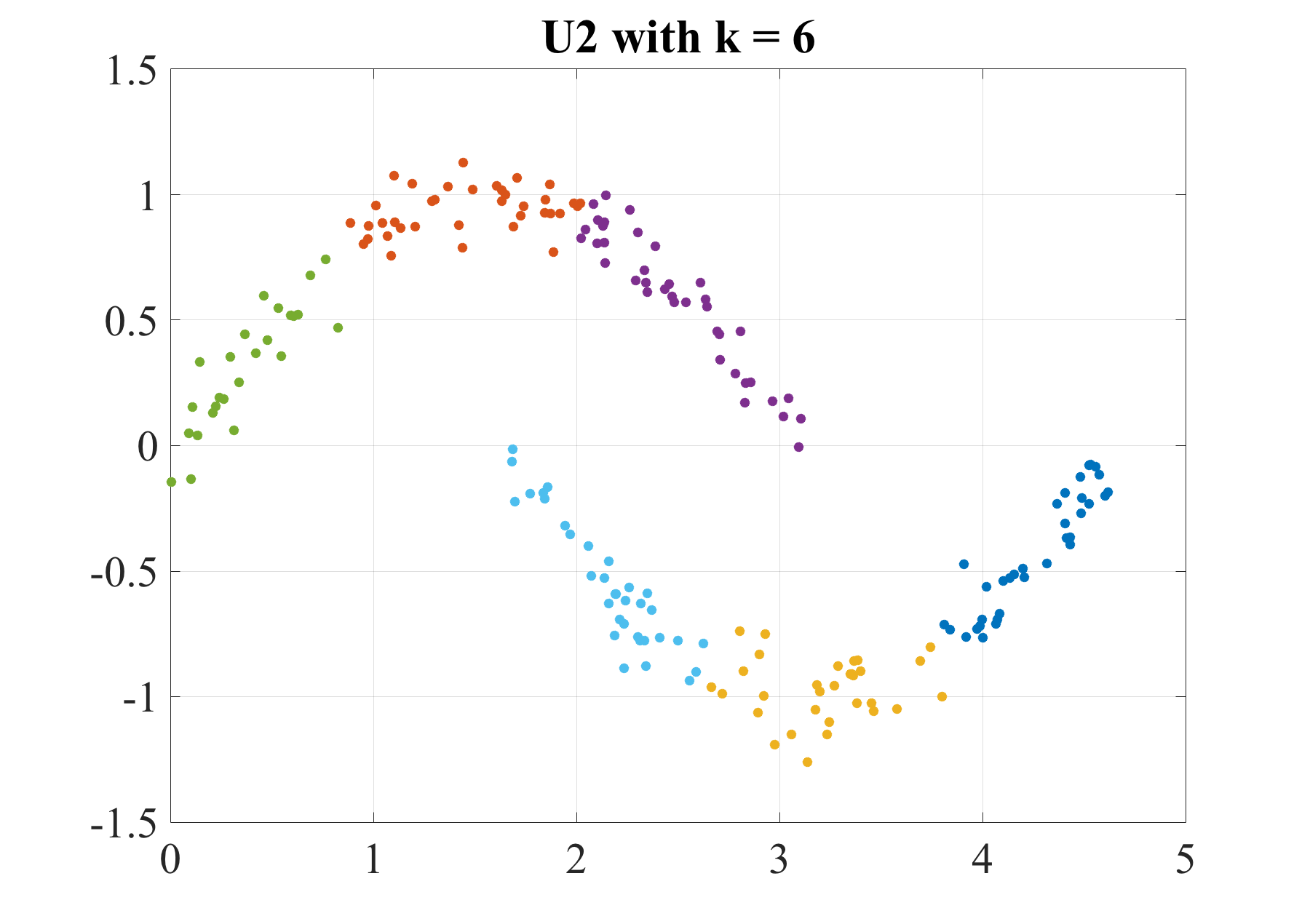}
\includegraphics[width=0.48\textwidth]{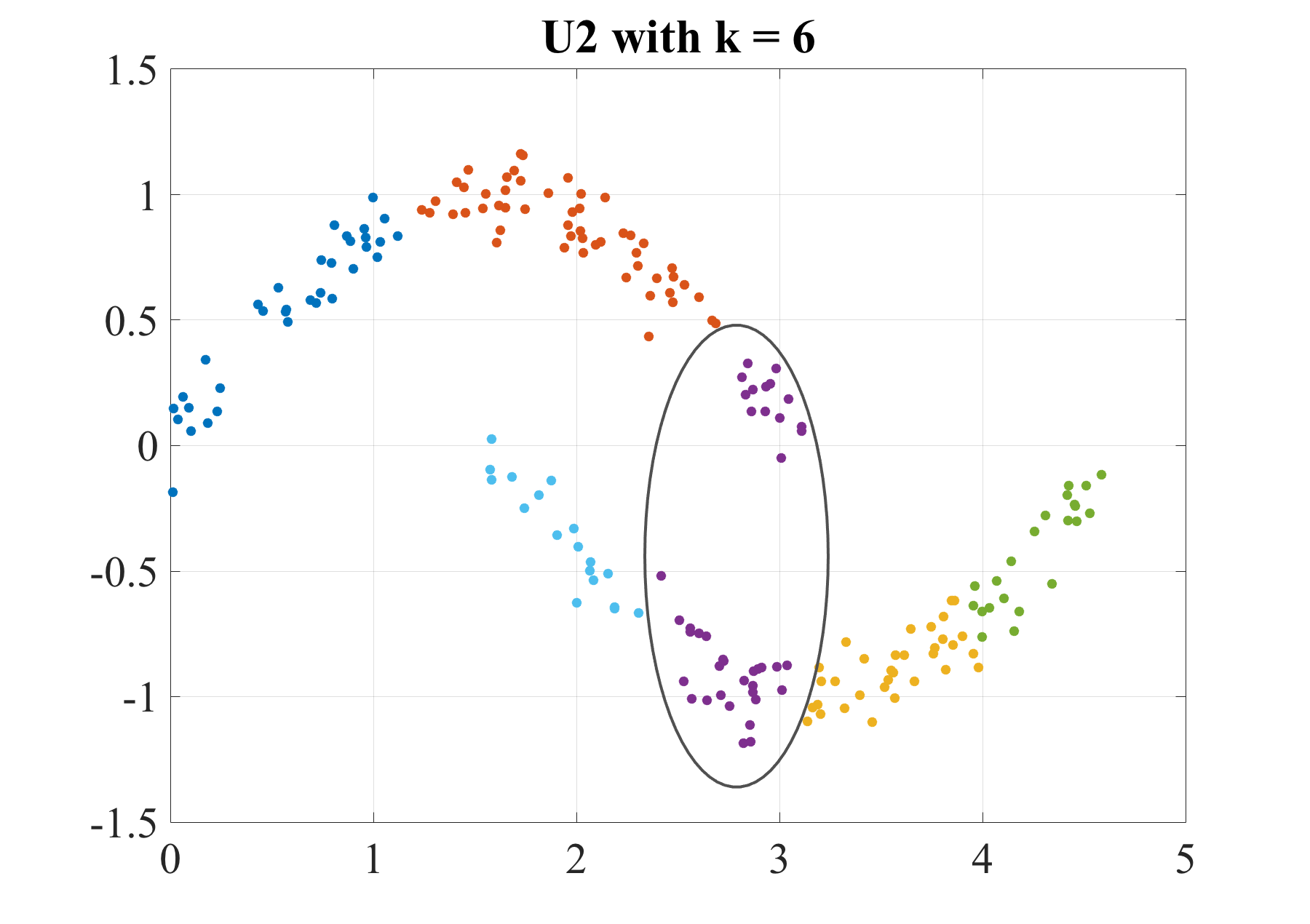}
\caption{Examples of partitions obtained by the k-means algorithm for the U2 and U4 dataset. Top row shows typical results for k-means with $k=2$ on both datasets. Bottom row shows results of k-means with $k=6$ on U2 datasets. The bottom-left plot shows a partition that is a refinement of the U2 cluster structure, which is not the case for the bottom-right plot due to the cluster circumscribed by the ellipse.}
\label{u2u4.png}
\end{figure}

The situation is different for both U-Shape datasets. The k-means algorithm is not able to essentially discover the cluster structure of the U2 and U4 datasets. Nevertheless, homogeneity is at the highest level for $k=2$ for both datasets. While this result is desirable for the U2 dataset at the first glance, it is unsatisfactory for the U4 dataset. 
The result for the U2 dataset with $k=2$ is only desirable at the first glance for the following reason: Although k-means performed stable and only a small fraction of partitions need to be removed in order to guarantee a unique mean partition, the discovered cluster structure does not properly match with the underlying cluster structure in the data as indicated by the top-left plot of Figure \ref{u2u4.png}. The same holds for the U4 dataset as shown by the top-right plot of Figure \ref{u2u4.png}.
Though the result for the U4 dataset with $k=2$ is unsatisfactory, closer inspection of the plots reveals that homogeneity has a moderate peak at $k=4$ for the U4 dataset not present for the U2 dataset. A further small peak is at $k=6$ for the U2 dataset. In the latter case, k-means frequently refines each of the two U-Shapes into three clusters (c.f.~plots at bottom row of Figure \ref{u2u4.png}). Such peaks may indicate that there could be a cluster structure but the underlying algorithm is unable to essentially discover this structure. A counter-example for this claim is the peak at $k=4$ for UD datasets as discussed in the previous experiment. The conclusion is that high homogeneity and stability are merely indicators for a possible cluster structure in the data but need further examination.

\subsection{Experiments on UCI Datasets}

\begin{table}[t]
\centering
\begin{tabular}{llrrr}
\hline
\hline
Dataset & Abbr. & Classes & Elements & Features \\
\hline
Banknote Authentication & bank & 2 & 1372 & 4\\
EEG Eye State& eye & 2 & 14980 & 14\\
Iris & iris & 3 & 150 & 4\\
Geographical Original of Music & music & 33 & 1059 & 68 \\
Pima Indians Diabetes & pima & 2 & 768 & 8\\
Connectionist Bench  & sonar & 2 & 208 & 60\\
\hline
\end{tabular}
\caption{Characteristics of six datasets from the UCI Machine Learning Repository.}
\label{tab:ucidata}
\end{table}

The goal of this experiment is to investigate how likely is a unique mean partition for real-world datasets. For this, we considered six datasets from the UCI Machine Learning Repository \cite{Lichman2013} listed in Table \ref{tab:ucidata}. For every dataset and for every $k\in \cbrace{2, \ldots, 10}$, we applied k-means $100$-times and recorded the $\alpha$-homogeneity.
 
 \medskip

\begin{figure}
\centering
\includegraphics[width=0.9\textwidth]{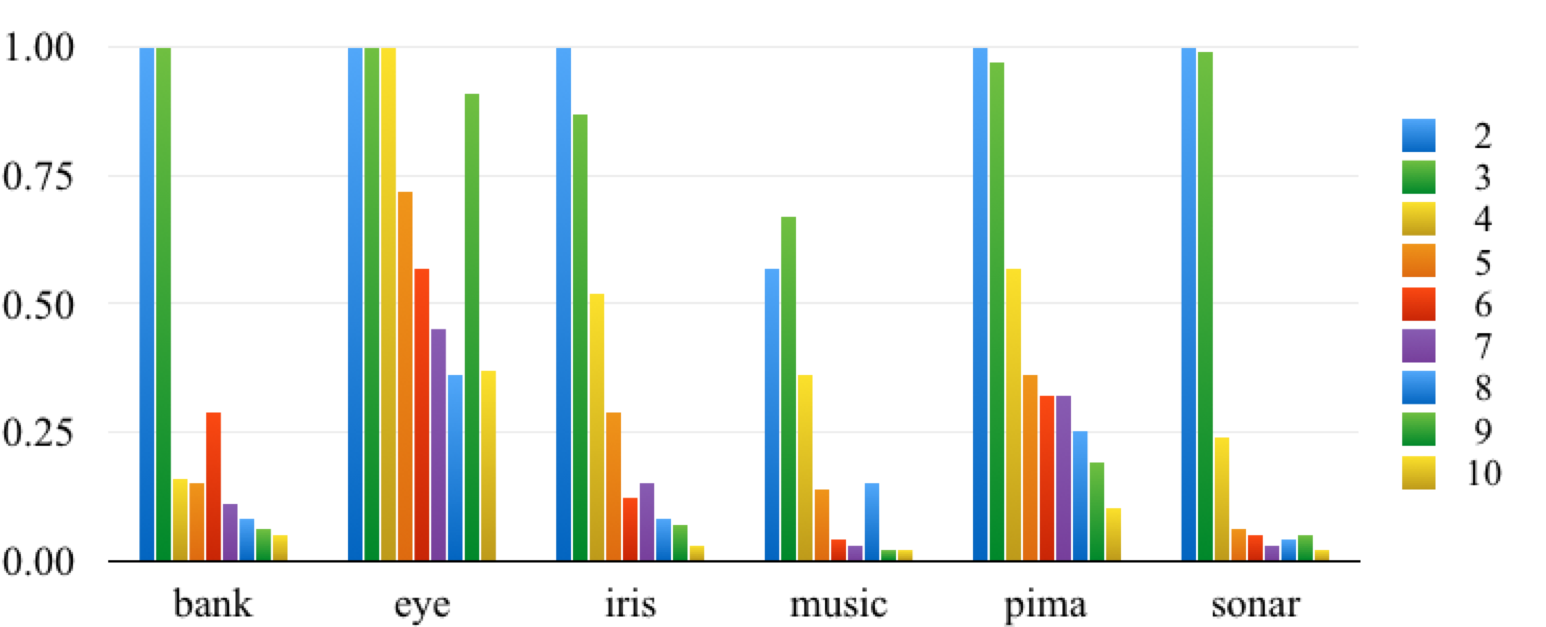}
\caption{$\alpha$-Homogeneities for each dataset as a function of the number $k$.}
\label{uci.png}
\end{figure}

Figure \ref{uci.png} shows the $\alpha$-homogeneities for each dataset as a function of the number $k$. We observed that (i) uniqueness of the mean partition is guaranteed for small values of $k$, and (ii) homogeneity decreases with increasing $k$. Exceptions from these general observations are the music and eye dataset.

Except for the music dataset, observation (i) shows that uniqueness of the mean partition can be guaranteed for real world data sets. This result indicates that uniqueness is of practical relevance and not a matter of exceptional cases. 

\begin{figure}
\centering
\includegraphics[width=0.5\textwidth]{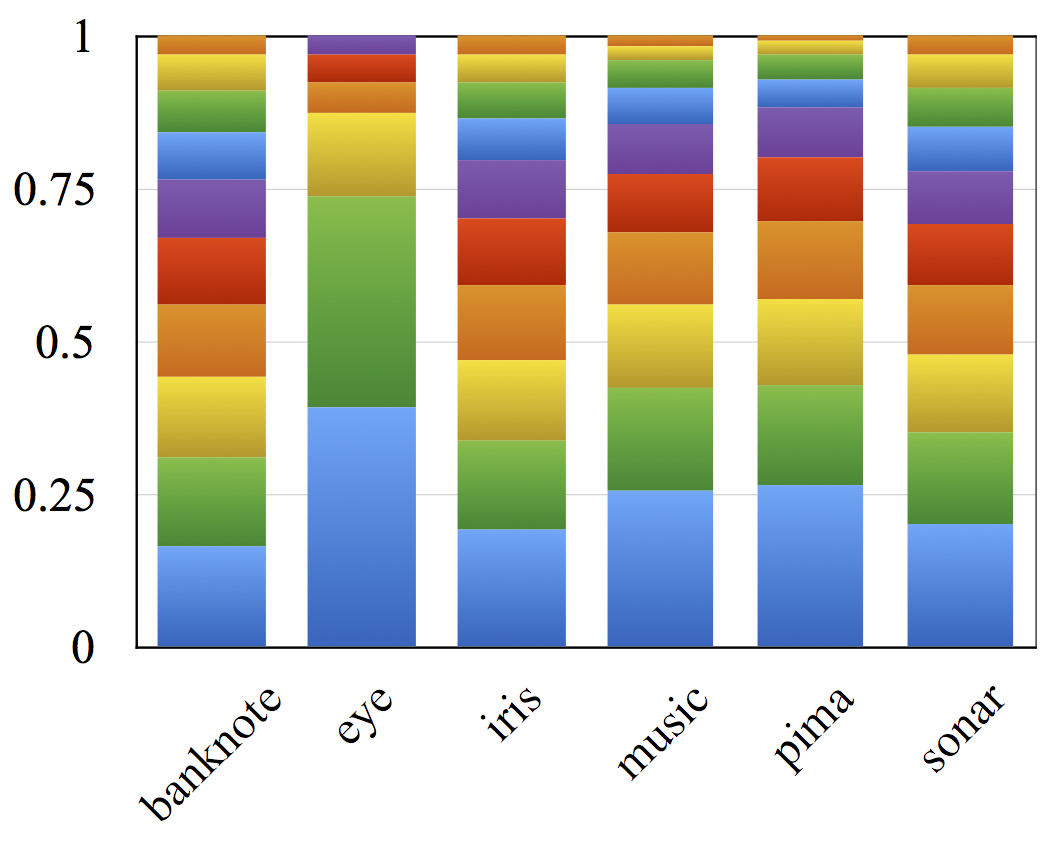}
\caption{Normalized cluster sizes for $k = 10$.}
\label{uci_err.png}
\end{figure}

Observation (ii) is in line with the results on synthetic data and an explanation follows the same argumentation as in Section \ref{subsubsec:exp:G4}. As shown in Figure \ref{uci_err.png}, the clusters are highly unbalanced for $k = 10$. Consequently, uniqueness of the mean partition is less likely for large $k$. Increasing the number $m$ of data points can improve homogeneity. This is possibly one reason why the eye dataset has substantially larger $\alpha$-homogeneity than the other datasets for all $k$ and has an additional peak of high homogeneity for $k = 9$, although it has the most imbalanced partitions for $k = 10$. 

\section{Conclusion}\label{sec:conclusion}

Uniqueness of the mean partition is a desirable property in consensus clustering that comes along with several benefits. We showed that both, the expected partition and the mean partition, are unique when the support is contained in an open subset of some max-hom ball. According to this condition, uniqueness is neither an exceptional nor a generic property. To cope with this issue, we proposed homogeneity as a measure of how close a sample is to having a unique mean. Homogeneity is not confined to consensus clustering but is also 
related to cluster stability. This in turn points to the possibility that cluster stability and diversity in consensus clustering can be conflicting goals. Homogeneity can be efficiently bounded from below by $\alpha$-homogeneity, which applies the degree of asymmetry of a partition. With $\alpha$-homogeneity, we can identify a sub-sample of the largest sub-sample of partitions that can be retained in order to guarantee a unique mean partition. Preliminary empirical results show that uniqueness occurs in real-world data and can be enforced by increasing the size of the dataset or by removing outlier partitions when $\alpha$-homogeneity is high. The results also indicate that $\alpha$-homogeneity can be used as a criterion for model selection. The results of this paper can be placed into the general context of a statistical theory of partitions, which embraces consensus clustering and cluster stability as special cases.

The main limitations of the proposed approach are twofold: restriction to the intrinsic metric on partitions derived from the Euclidean distance and restriction of uniqueness conditions on max-hom balls. Generalizing both restrictions are two possible directions of future research. Another important issue is to understand when cluster stability and diversity in consensus clustering collide.


\begin{appendix}
\small

\section{Preliminaries}

This section presents technicalities useful for proving the results proposed in the main text. 

\subsection{Notations}\label{subsec:notations}
We use the following notations: By $\overline{\S{U}}$ we denote the closure of a subset $\S{U} \subseteq \S{X}$, by $\partial \S{U}$ the boundary of $\S{U}$, and by $\S{U}^\circ$ the open subset $\overline{\S{U}} \setminus \partial \S{U}$. The action of permutation $\vec{P} \in \Pi$ on the subset $\S{U}\subseteq \S{X}$ is the set defined by $\vec{P}\,\S{U} = \cbrace{\vec{PX} \, :\, \vec{X} \in \S{U}}$. A transposition is a permutation matrix $\vec{P} \in \Pi$ that permutes exactly two rows. A basic result from algebra is that any permutation matrix $\vec{P} = \Pi^*$ can be written as a matrix product $\vec{P} = \vec{Q}_1 \cdots \vec{Q}_t$ of transpositions $\vec{Q}_i \in \Pi$ with minimum number $t > 0$ of factors.

\subsection{Dirichlet Fundamental Domains}
A subset $\S{F}$ of $\S{X}$ is a fundamental set for $\Pi$ if and only if $\S{F}$ contains exactly one representation $\vec{X}$ from each orbit $\bracket{\vec{X}} \in \S{X}/\Pi$. 
A fundamental domain of $\Pi$ in $\S{X}$ is a closed set $\S{F} \subseteq \S{X}$ that satisfies 
\begin{enumerate}
\item $\displaystyle\S{X} = \bigcup_{\vec{P} \in \Pi} \vec{P}\S{F}$
\item $\vec{P} \S{F}^\circ \cap \S{F}^\circ = \emptyset$ for all $\vec{P} \in \Pi^*$.
\end{enumerate}

\begin{proposition}
Let $\vec{Z}$ be a representation of an asymmetric partition $Z \in \S{P}$. Then 
\[
\S{D}_{\vec{Z}} = \cbrace{\vec{X} \in \S{X} \,:\, \norm{\vec{X} - \vec{Z}} \leq \norm{\vec{X} - \vec{PZ}} \text{ for all }\vec{P} \in \Pi}
\]
is a fundamental domain, called \emph{Dirichlet fundamental domain} centered at $\vec{Z}$. 
\end{proposition}
\noindent
\proof \cite{Ratcliffe2006}, Theorem 6.6.13. \qed

\medskip

\noindent
The next result list some properties of Dirichlet fundamental domains. 

\begin{proposition}\label{proof:properties-of-DFD}
Let $\S{D}_{\vec{Z}}$ be a Dirichlet fundamental domain centered at representation $\vec{Z}$ of an asymmetric partition $Z \in \S{P}$. Then the following properties hold:
\begin{enumerate}
\item There is a fundamental set $\S{F}_{\vec{Z}}$ such that $\S{D}_{\vec{Z}}^\circ \subseteq \S{F}_{\vec{Z}} \subseteq \S{D}_{\vec{Z}}$.
\item We have $\vec{Z} \in \S{D}_{\vec{Z}}^\circ$.
\item Every point $\vec{X} \in \S{D}_{\vec{Z}}^\circ$ represents an asymmetric partition.
\item Suppose that $\vec{X}, \vec{PX} \in \S{D}_{\vec{Z}}$ for some $\vec{P} \in \Pi^*$. Then $\vec{X}, \vec{PX} \in \partial \S{D}_{\vec{Z}}$.
\item $\vec{P} \S{D}_{\vec{Z}} = \S{D}_{\vec{PZ}}$ for all $\vec{P} \in \Pi$.
\end{enumerate}
\end{proposition}

\noindent
\proof \ 
The proof follows \cite{Jain2015}, Prop.~3.13 but is adapted to the notation and terminology of this contribution.
\setcounter{part_counter}{0}
\begin{part}
\cite{Ratcliffe2006}, Theorem 6.6.11.
\end{part}

\begin{part}
Since $Z$ is asymmetric, we have $\norm{\vec{Z} - \vec{Z}} < \norm{\vec{Z} - \vec{PZ}}$ for all $\vec{P} \in \Pi^*$. This shows that $\vec{Z} \in \S{D}_{\vec{Z}}^\circ$.
\end{part}

\begin{part}
Let $\vec{X} \in \S{D}_{\vec{Z}}^\circ$ be a representation of partition $X$. Suppose that $X$ is symmetric. Then there is a $\vec{P} \in \Pi^*$ with $\vec{X} = \vec{PX}$. This implies $\vec{X} \in \vec{P}\, \S{D}_{\vec{Z}} \cap \S{D}_{\vec{Z}}$. Then $\vec{X} \in \partial\S{D}_{\vec{Z}}$ is a boundary point of $\S{D}_{\vec{Z}}$ by \cite{Ratcliffe2006}, Theorem 6.6.4. This contradicts our assumption that $\vec{X} \in \S{D}_{\vec{Z}}^\circ$ and shows that $\vec{X}$ is asymmetric.
\end{part}

\begin{part}
From $\vec{X}, \vec{PX} \in \S{D}_{\vec{Z}}$ follows $\norm{\vec{X} - \vec{Z}} = \norm{\vec{PX} - \vec{Z}}$. Since $\Pi$ acts by isometries, we have $\norm{\vec{X} - \vec{Z}} = \norm{\vec{PX} - \vec{PZ}}$. Thus, we have $\norm{\vec{PX} - \vec{Z}} = \norm{\vec{PX} - \vec{PZ}}$. This shows that $\vec{PX} \in \partial \S{D}_{\vec{Z}}$. Let $\vec{P}' \in \Pi$ be the inverse of $\vec{P}$. Since $\vec{P} \neq \vec{I}$, we have $\vec{P}' \neq \vec{I}$. Then
\begin{align*}
\norm{\vec{X} - \vec{Z}} &=\norm{\vec{PX} - \vec{Z}} 
= \norm{\vec{P'PX} - \vec{P'Z}}
= \norm{\vec{X} - \vec{P'Z}},
\end{align*}
From $\norm{\vec{X} - \vec{Z}} = \norm{\vec{X} - \vec{P'Z}}$ follows $\vec{X} \in \partial \S{D}_{\vec{Z}}$.
\end{part}

\begin{part}
Let $\vec{X} \in \vec{P}\, \S{D}_{\vec{Z}}$. We have $\norm{\vec{X} - \vec{PZ}} \leq \norm{\vec{P'X} - \vec{PZ}} $ for all $\vec{P}' \in \Pi$ showing that $\vec{X} \in \S{D}_{\vec{PZ}}$. Now assume that $\vec{X} \in \S{D}_{\vec{PZ}}$. Let $\vec{P}' \in \Pi$ be the inverse of $\vec{P}$. Then we have 
\[
\norm{\vec{X} - \vec{PZ}} = \norm{\vec{P'X} - \vec{P'PZ}} = \norm{\vec{P'X} - \vec{Z}} 
\] 
by isometry of $\vec{P}'$. Hence, $\vec{P'X} \in \S{D}_{\vec{Z}}$ and therefore $\vec{PP'X} = \vec{X} \in \vec{P}\, \S{D}_{\vec{Z}}$.
\end{part}
\qed

\subsection{Cross Sections}

Suppose that $\S{D}_{\vec{Z}}$ is the Dirichlet fundamental domain centered at representation $\vec{Z}$ of an asymmetric partition $Z \in \S{P}$. A map $\mu:\S{P} \rightarrow \S{D}_{\vec{Z}}$ is a cross section into $\S{D}_{\vec{Z}}$, if $\pi(\mu(X)) = X$ for all partitions $X \in \S{P}$. 

\begin{proposition}\label{prop:cross-section}
Let $\mu:\S{P} \rightarrow \S{D}_{\vec{Z}}$ be a cross section into a Dirichlet fundamental domain $\S{D}_{\vec{Z}}$ centered at representation $\vec{Z}$ of an asymmetric partition $Z \in \S{P}$. Then the following properties hold:
\begin{enumerate}
\item $\mu$ is injective.
\item $\mu(\S{P})$ is a fundamental set. 
\item $\mu$ is a measurable mapping.
\end{enumerate}
\end{proposition}

\noindent
\proof Both assertions directly follow from the definitions of cross section and fundamental set. max-hom of $\mu$ directly follows from the property $\pi\circ \mu = \id$. Again from $\pi\circ \mu = \id$ follows that $\mu$ maps partitions to representations. Finally, since $\mu$ is injective, the image $\mu(\S{P})$ contains exactly one representation of each partition. Hence, $\mu(\S{P})$ is a fundamental set. Finally, $\mu$ is measurable, because $\mu^{-1} = \pi$ and $\pi$ is an open mapping. 
\qed

\medskip

Let $\args{\S{P}, \S{B}, Q}$ be a measurable space. A cross section $\mu:\S{P} \rightarrow \S{D}_{\vec{Z}}$ is a measurable map that gives rise to a measurable space $\args{\S{D}_{\vec{Z}}, \S{B}_\mu, q}$.

\section{Proofs}

\subsection{Proof of Theorem \ref{theorem:uniqueness-of-mean}}

Parts 1--4 show uniqueness of the expected partition and Part 5 shows uniqueness of the mean partition. 

\setcounter{part_counter}{0}
\begin{part}
Both assertions trivially hold for asymmetric partitions $Z$, because $\S{S}_Q \subseteq \S{B}_Z = \cbrace{Z}$.
\end{part}

\begin{part}
Let $Z \in \S{P}$ be an asymmetric partition such that $\S{S}_Q \subseteq \S{B}_Z$. We select an arbitrary representation $\vec{Z} \in Z$ and an arbitrary cross section $\mu:\S{P} \rightarrow \S{D}_{\vec{Z}}$. Let $\S{S}_{\vec{Z}} = \mu\args{\S{S}_Q}$ be the image of the support $\S{S}_Q$. Since $\S{B}_Z$ is a homogeneous ball, we have
\begin{align*}
F(Y) &= \int_{\S{P}} \delta(X, Y)^2 dQ(X) = \int_{\S{S}_Q} \delta(X, Y)^2 dQ(X)= \int_{\S{S}_{\vec{Z}}} \normS{\mu(X)-\mu(Y)}{^2} d q(\mu(X)), 
\end{align*}
where $q$ is the image measure of measure $Q$ under cross section $\mu$. The function
\[
f(\vec{Y}) = \int_{\S{S}_{\vec{Z}}} \normS{\vec{X}-\vec{Y}}{^2} d q(\vec{X})
\]
has a unique minimum $\vec{M} \in \S{X}$ representing partition $M \in \S{P}$. From 
\[
\S{S}_{\vec{Z}} \subseteq \S{B}(\vec{Z}, \alpha_Z/4) \subsetneq \S{D}^\circ_{\vec{Z}}
\] 
together with Prop.~\ref{prop:cross-section} follows that partition $M$ is independent of the choice of a particular cross section along $\vec{Z}$. From Prop.~\ref{proof:properties-of-DFD} follows that partition $M$ is independent of the choice of a particular representation of $Z$. It remains to show that partition $M$ of the second part of this proof is independent of the choice of partition $Z$ that satisfies $\S{S}_Q \subseteq \S{B}_Z$. This is proved in the sequel. 
\end{part}

\begin{part}
Suppose that $Z' \in \S{P}$ is a partition satisfying $\S{S}_Q \subseteq \S{B}_{Z'}$. Let $\vec{Z}' \in Z'$ be a representation such that $\vec{Z}, \vec{Z}' \in \S{D}_{\vec{Z}} \cap \S{D}_{\vec{Z}'}$ and let $\mu':\S{P} \rightarrow \S{D}_{\vec{Z}'}$ be a cross section. By $\S{S} = \mu\args{\S{S}_Q}$ and $\S{S}' = \mu'\args{\S{S}_Q}$ we denote the images of $\S{S}_Q$ under the cross sections $\mu$ and $\mu'$, respectively. 

We show that there is a permutation matrix $\vec{P} \in \Pi$ such that $ \S{S}' = \vec{P} \S{S}$. Observe that the composition $\mu' \circ \pi: \S{S} \rightarrow \S{S}'$ is bijective. Let $\vec{X} \in \S{S}$ and $\vec{X}' \in \S{S}'$ be representations such that $\vec{X}' = \mu'(\pi(\vec{X}))$. Since $\mu'$ is a cross section, both elements $\vec{X}$ and $\vec{X}'$ represent the same partition $\pi(\vec{X}) \in \S{S}_Q$. Then there is a permutation matrix $\vec{P} \in \Pi$ such that $\vec{X}' = \vec{PX}$. We assume that there are representations $\vec{Y} \in \S{S}$ and $\vec{Y}' \in \S{S}'$ such that 
\[
\vec{Y}' = \mu'(\pi(\vec{Y})) \neq \vec{PY}.
\]
Since $\vec{Y}$ and $\vec{Y}'$ represent the same partition $\pi(\vec{Y}) \in \S{S}_Q$, there is another permutation matrix $\vec{Q} \in \Pi\setminus\cbrace{\vec{P}}$ such that $\vec{Y}' = \vec{QY}$. We find that $\vec{PY} \neq \vec{QY}$. To see this, observe that $\vec{Y} \in \S{S} \subset \S{D}_{\vec{Z}}^\circ$ is an interior point of $\S{D}_{\vec{Z}}$. From Prop.~\ref{proof:properties-of-DFD} follows that $Y = \pi(\vec{Y})$ is asymmetric. This implies that $\vec{Y} \neq \vec{RY}$ for all $\vec{R} \in \Pi^*$. Since $\vec{P} \neq \vec{Q}$, we obtain $\vec{PY} \neq \vec{QY}$.

Although $\vec{PY} \neq \vec{QY}$, we have
\begin{align}\label{eq:theorem:uniqueness-of-mean:part3:case2:01}
\norm{\vec{X} - \vec{Y}} = \norm{\vec{PX} - \vec{PY}} = \norm{\vec{PX} - \vec{QY}}.
\end{align}
The first equation holds, because $\Pi$ acts isometrically on $\S{X}$. The second equation follows from $\vec{X}', \vec{Y}' \in \S{B}_{\vec{Z}'}$ together with the fact that $\S{B}_{\vec{Z}'}$ is a homogeneous ball. From Equation \eqref{eq:theorem:uniqueness-of-mean:part3:case2:01} follows that the Dirichlet fundamental domain $\S{D}_{\vec{X}'}$ centered at $\vec{X}' = \vec{PX}$ contains $\vec{PY}$ and $\vec{QY}$. Then by Prop.~\ref{proof:properties-of-DFD} both representations $\vec{PY}$ and $\vec{QY}$ are elements of the boundary of $\S{D}_{\vec{X}'}$. By assumption, only $\vec{QY} \in \S{S}'$ is an element of $\S{B}_{\vec{Z}'}$. Moreover, since $ \S{S}' \subsetneq \S{B}_{\vec{Z'}}^\circ$, we have 
\[
\S{V} = \S{D}_{\vec{QX}}^\circ \cap \S{B}_{\vec{Z'}} \neq \emptyset.
\] 
Suppose that $\vec{V} \in \S{V}$ is a representation of partition $V$. Then $\delta(X, V) < \norm{\vec{X}' - \vec{V}}$, because $\vec{V}$ is in the interior of $\S{D}_{\vec{QX}}$ and $\vec{X}'$ is in the interior of $\S{D}_{\vec{X'}}$. This contradicts the assumption that $\S{B}_{Z'}$ is a homogeneous ball isometric to $\S{B}_{\vec{Z}'}$. Hence, $\vec{Y}' = \vec{PY}$ and therefore $ \S{S}' = \vec{P} \S{S}$. 
\end{part}

\begin{part}
This part shows that partition $M$ is independent of the choice of $Z$. A permutation matrix $\vec{P} \in \Pi$ gives rise to a diffeomorphism 
\[
L_{\vec{P}}: \S{X} \rightarrow \S{X}, \quad \vec{X} \mapsto \vec{P}^{-1}\vec{X}
\]
with Jacobi matrix $L_{\vec{P}}' = \vec{P}^{-1}$ and $\abs{\det L_{\vec{P}}'} = \abs{\det \vec{P}^{-1}} = 1$.
Let $q'$ be the image measure of $Q$ under the cross section $\mu'$. From 
\[
\mu'(\S{S}_Q) = \S{S}' = \vec{P}\S{S} = L_{\vec{P}}^{-1}\circ\mu(\S{S}_Q)
\]
follows $\mu = L_{\vec{P}} \circ \mu'$. Then we have 
\[
q = \mu(Q) = L_{\vec{P}} \circ \mu' (Q) = L_{\vec{P}}(q'),
\]
that is $q$ is the image measure of $q'$ under $L_{\vec{P}}$. For every $\vec{Y} \in \S{S}$, we define the function 
\[
g_{\vec{Y}}(\vec{X}) = \normS{\vec{X} - \vec{Y}}{^2}.
\]
Then we rewrite $f(\vec{Y})$ by
\[
f(\vec{Y}) = \int_{\S{S}} \normS{\vec{X}-\vec{Y}}{^2} d q(\vec{X})=\int_{\S{S}} g_{\vec{Y}}(\vec{X})dq(\vec{X}). 
\]
Applying the transformation formula for integrals gives
\begin{align*}
f(\vec{Y})
=\int_{\S{S}} g_{\vec{Y}}(\vec{X})dq(\vec{X}) 
= \int_{\S{S}'} g_{\vec{Y}} \circ L_{\vec{P}}(\vec{X}') \cdot\abs{\det L_{\vec{P}}'}dq'(\vec{X}') 
= \int_{\S{S}'} \normS{\vec{P}^{-1}\vec{X}'-\vec{Y}}{^2} d q'(\vec{X}').
\end{align*}
Since $\Pi$ acts isometrically on $\S{X}$, we have
\[
f(\vec{Y}) = \int_{\S{S}'} \normS{\vec{P}\vec{P}^{-1}\vec{X}'-\vec{P}\vec{Y}}{^2} d q'(\vec{X}') = \int_{\S{S}'} \normS{\vec{X}'-\vec{P}\vec{Y}}{^2} d q'(\vec{X}') = f'(\vec{PY}).
\]
From $\vec{Y} \in \S{S}$ and $\S{S}' = \vec{P}\S{S}$ follows $\vec{PY} \in \S{S}'$. Then the unique minimizer $\vec{M}$ of $f(\vec{Y})$ gives $\vec{PM}$ as the unique minimizer of $f'(\vec{Y}')$ on $\S{S}'$. Both elements $\vec{M}$ and $\vec{PM}$ represent the same partition $M$. This shows that $M$ is independent of the choice of $Z$ such that $\S{S}_Q \subseteq \S{B}_Z$. Hence, $M$ is the unique minimizer of $F(Z)$. 
\end{part}

\begin{part}
We show uniqueness of the mean partition. Let $\S{S}_n = \args{X_1, \ldots, X_n}$ be a sample of $n$ partitions. Let $\S{S}_{\cbrace{n}} = \cbrace{X_1, \ldots, X_n}$ denote the 
the set of partitions induced by $\S{S}_n$. We define a probability measure $Q_n$ as a probability mass function of the form
\[
Q_n(X) = \frac{1}{n} \sum_{i=1}^n \delta_{X, X_i}
\]
for all $X \in \S{S_Q}$, where $\delta_{X,Y}$ is the Kronecker delta that gives one when $X = Y$ agree and zero, otherwise. We use $Q_n$ as probability measure and $\S{S}_Q$ as set containing the support of $Q_n$. Then the assertion follows from Part 1--4 of this proof.
\end{part}


\subsection{Proof of Prop.~\ref{prop:asymmetry-max-hom} }

\setcounter{part_counter}{0}
\begin{part}
The first assertion holds for symmetric partitions $Z$, because $\alpha_Z = 0$ and therefore $\alpha_Z/4 \leq \rho_Z$. We assume that $Z$ is asymmetric. The group $\Pi$ is a discontinuous group acting isometrically on $\S{X}$. The isotropy group 
\[
\Pi_{\vec{Z}} = \cbrace{\vec{P} \in \Pi \,:\, \vec{P}\vec{Z} = \vec{Z}}
\]
is trivial for any representation $\vec{Z} \in Z$. Since $\S{P} \cong \S{X}/\Pi$, we have a bijective isometry 
\[
\phi: \S{B}\!\args{\vec{Z}, \rho} \longrightarrow \S{B}\!\args{Z, \rho}, \quad \vec{X} \mapsto \pi(\vec{X}).
\]
for all $0 < \rho \leq \alpha_Z/4$ by \cite{Ratcliffe2006}, Theorem 13.1.1. Setting $\psi = \phi^{-1}$ we find that $\S{B}(Z, \rho)$ is a homogeneous ball. This shows $\alpha_Z/4 \leq \rho_Z$.
\end{part}

\begin{part}
We show the second assertion. From Part 1 of this proof follows $\alpha_Z > 0 \, \Rightarrow \, \rho_Z > 0$. We show the opposite direction. Let $\rho_Z > 0$. We assume that $\alpha_Z = 0$. Suppose that $\vec{Z} \in Z$ is a representation. Then there is a permutation matrix $\vec{P} \in \Pi^*$ such that $\vec{Z} = \vec{PZ}$. Consider the ball $\S{B}_{\varepsilon} = \S{B}(\vec{Z}, \varepsilon)$ for $\varepsilon > 0$. Suppose that $\vec{X} \in \S{B}_{\varepsilon}$ is an element representing an asymmetric partition. Such am element exists according to Prop.~\ref{prop:properties:asymmetry}(1). Then $\vec{X} \neq \vec{PX}$ and $\vec{X}, \vec{PX} \in \S{B}_{\varepsilon}$. This shows that there is no bijective mapping between $\S{B}(Z, \varepsilon)$ and $\S{B}_{\varepsilon}$ for any $\varepsilon > 0$. This contradicts our assumption that $\rho_Z > 0$. Hence, we have $\alpha_Z > 0$. 
\end{part}

\subsection{Proof of Prop.~\ref{prop:properties:asymmetry}}
\setcounter{part_counter}{0}
\begin{part}
To prove the first assertion, it is sufficient to show that the set of asymmetric partitions forms an open and dense subset in $\S{P}$. The projection $\pi:\S{X} \rightarrow \S{X}/\Pi$ is open and surjective. Then the image $\pi(\S{U})$ of an open and dense subset $\S{U} \subseteq\S{X}$ is open and dense in $\S{X}/\Pi$. To see this observe that from $\pi(\S{X}) = \pi(\overline{\S{U}})$ and surjectivity of $\pi$ follows $ \pi(\overline{\S{U}}) = \S{X}/\Pi$. From $\S{X}/\Pi = \pi(\overline{\S{U}})\subseteq \overline{\pi(\S{U})}$ follows that $\pi(\S{U})$ is open and dense in $\S{X} /\Pi$. 

Now let $\vec{Z} \in \S{X}$ be a representation of an asymmetric partition $Z \in \S{P}$. Suppose that $\mu:\S{P} \rightarrow \S{D}_{\vec{Z}}$ is a cross section. From Prop.~\ref{prop:cross-section} follows that $\mu(\S{P}) \subset \S{D}_{\vec{Z}}$ is a fundamental set containing the open set $\S{D}_{\vec{Z}}^\circ$. Since $\S{D}_{\vec{Z}}^\circ$ is open and dense in $\S{D}_{\vec{Z}}$, we find that $\pi\args{\S{D}_{\vec{Z}}^\circ}$ is open and dense in $\S{P}$. This shows that almost all partitions are asymmetric.
\end{part}

\begin{part}
We show the second assertion. Let $Z\in \S{P}$ be an asymmetric partition. Suppose that $\vec{Z} \in Z$ is a representation matrix with rows $\vec{z}_1, \ldots. \vec{z}_\ell$. We assume that $Z$ has two identical clusters. Then there are two distinct indices $1 \leq p < q \leq \ell$ such that $\vec{z}_p = \vec{z}_q$. Let $\vec{P} \in \Pi^*$ be the permutation matrix that swaps rows $p$ and $q$. Then we have $\alpha_Z \leq \norm{\vec{Z} -\vec{PZ}} = 0$. This contradicts our assumption that $Z$ is asymmetric. Hence, the clusters of $Z$ are mutually distinct. 

Next, we assume that all clusters of $Z$ are mutually distinct. Suppose that $Z$ is asymmetric. Then there is a representation $\vec{Z} \in Z$ and a permutation matrix $\vec{P} \in \Pi^*$ such that $\alpha_Z = \norm{\vec{Z}-\vec{PZ}}$. As stated in Section \ref{subsec:notations}, we can express $\vec{P}$ as a minimal product of $t > 0$ transpositions. Hence, representation $\vec{Z}$ has at least one pair of identical rows. This contradicts our assumption that the clusters of $Z$ are mutually distinct showing that partition $Z$ is asymmetric. 
\end{part}

\subsection{Proof of Prop.~\ref{prop:properties:alpha}}

The first three parts of this proof prepare the proofs of the assertions shown in Part 4--6. 

\setcounter{part_counter}{0}
\begin{part}
Let $\vec{P} \in \Pi^*$ be a transposition that permutes rows $p < q$. Then we have 
\begin{align*}
\normS{\vec{Z} - \vec{PZ}}{^2} = \normS{\vec{z}_p-\vec{z}_q}{^2} + \normS{\vec{z}_q-\vec{z}_p}{^2} = 2 \normS{\vec{z}_p-\vec{z}_q}{^2}. 
\end{align*}
This gives $\norm{\vec{Z} - \vec{PZ}} = \sqrt{2} \norm{\vec{z}_p-\vec{z}_q}$.
\end{part}

\begin{part}
Let $\vec{P}, \vec{Q} \in \Pi^*$ be two different transpositions. Suppose that $\vec{P}$ permutes rows $p < q$ and $\vec{Q}$ permutes rows $r < s$ such that either (i) $\cbrace{p, q} \cap \cbrace{r, s} = \emptyset$ or (ii) $q = r$. For case (i), we have
\begin{align*}
\normS{\vec{Z} - \vec{PQZ}}{^2} = 2\normS{\vec{z}_p-\vec{z}_q}{^2} + \;2\normS{\vec{z}_r-\vec{z}_s}{^2}
\end{align*}
according to the first part of this proof. This implies $\norm{\vec{Z} - \vec{PZ}} \leq \norm{\vec{Z} - \vec{PQZ}}$ and $\norm{\vec{Z} - \vec{QZ}} \leq \norm{\vec{Z} - \vec{PQZ}}$. 
\end{part}

\begin{part}\label{part:prop:properties:alpha:01}
As stated in Section \ref{subsec:notations}, we can write $\vec{P} = \Pi^*$ as a matrix product $\vec{P} = \vec{Q}_1 \cdots \vec{Q}_t$ of transpositions $\vec{Q}_i \in \Pi$ with minimum number $t > 0$ of factors. From the second part of this proof follows
\begin{align*}
\norm{\vec{Z} - \vec{Q}_i\vec{Z}} &\leq \norm{\vec{Z} - \vec{Q}_1 \vec{Q}_2 \cdots \vec{Q}_t \vec{Z}} = \norm{\vec{Z} - \vec{PZ}} 
\end{align*}
for all $i \in \cbrace{1, \ldots, t}$. This shows that it is sufficient to restrict to transpositions for determining the degree of asymmetry of a partition.
\end{part}

\begin{part}
Let $\Pi_\tau$ denote the subset of all transpositions. From part 1 and \ref{part:prop:properties:alpha:01} of this proof follows
\begin{align*}
\alpha_Z &= \min \cbrace{\norm{\vec{Z} - \vec{P}\vec{Z}} \,:\, \vec{P} \in \Pi_\tau}
= \min\cbrace{\sqrt{2}\norm{\vec{z}_p - \vec{z}_q} \,:\, 1 \leq p < q \leq \ell }.
\end{align*}
This shows the first assertion.
\end{part}

\begin{part}
The second assertion assumes that $Z$ is a hard partition. Then the elements of $\vec{Z}$ take binary values from $\cbrace{0,1}$ such that 
$\vec{z}_p^T\vec{z}_q = 0$ for all $1 \leq p < q \leq \ell$. Moreover, $n_p = \vec{z}_k^T\vec{z}_k$ is the size of the $k$-th cluster. Then we have
\[
\normS{\vec{z}_p - \vec{z}_q}{^2} = \vec{z}_p^T\vec{z}_p - 2 \vec{z}_p^T\vec{z}_q + \vec{z}_q^T\vec{z}_q = n_p + n_q.
\]
From the fourth part of this proof follows
\[
\alpha_Z = \min\cbrace{\sqrt{2 \args{n_p + n_q}} \,:\, 1 \leq p < q \leq \ell }.
\]
This implies the second assertion.
\end{part}

\begin{part}
The second assertion assumes that $Z$ is an asymmetric hard partition. We first prove the lower bound of $\alpha_Z$. From Prop.~\ref{prop:properties:asymmetry} follows that $Z$ has at most one empty cluster. Then $m_1 = 0$ and $m_2 > 0$. Using part 5 of this proof, we obtain $\sqrt{2} \leq \sqrt{2m_2} = \alpha_Z$. Next, we show the upper bound of $\alpha_Z$. From the strong form of the pigeonhole principle follows that there is a cluster with at least $\mu_1 = \lceil m/\ell \rceil$ elements. Let $m' = m - \mu_1$ be the number of remaining elements. Then again applying the pigeonhole principle gives a cluster with at least $\mu_2 = \lceil m'/(\ell-1) \rceil$ elements. From $\mu_1 \geq m/\ell$ follows
\[
\mu_2 = \frac{m - \mu_1}{\ell - 1} \leq \frac{m - m/\ell}{\ell - 1} = \frac{m}{\ell} \leq \left\lceil\frac{m}{\ell}\right\rceil.
\]
We can bound the cardinality of the two smallest clusters by 
\[
m_1+m_2 \leq \mu_1 + \mu_2 \leq 2 \left\lceil\frac{m}{\ell}\right\rceil.
\]
Using part 5 of this proof shows the third assertion.
\end{part}

\subsection{Proof of Equation \eqref{eq:F<I}}\label{proof:eq:F<I}
We have
\begin{align*}
I_{n, k} 
&= \frac{1}{n^2} \sum_{i, j}\Delta_k\!\args{X_i, X_j}\\
&\geq \frac{1}{n^2} \sum_{i,j}\Delta_k\!\args{M_k, X_j}\\
&= \frac{1}{n} \sum_{j}\Delta_k\!\args{M_k, X_j}\\
&= F_{n,k}\!\args{M_k}.
\end{align*}
The inequality in the second line holds, because $M_k$ is a mean or medoid. This shows the assertion.

\end{appendix}

\end{document}